\documentclass[preprint,12pt,authoryear]{elsarticle}

\usepackage{amssymb}
\usepackage{amsmath}
\usepackage{array}
\usepackage{graphicx}
\usepackage{multirow}
\usepackage{ragged2e}
\usepackage{colortbl}
\usepackage{subcaption}
\usepackage{wrapfig}
\usepackage{float}
\usepackage{tabularray}
\usepackage{xurl}
\usepackage[hidelinks]{hyperref}
\usepackage{bm}
\usepackage{mathtools}
\usepackage{xcolor}

\newcommand{\norm}[1]{\left\lVert #1 \right\rVert}
\newcommand{\sg}{\operatorname{sg}}
\newcommand{\OneHot}{\operatorname{OneHot}}
\newtheorem{theorem}{Theorem}
\definecolor{darkgreen}{RGB}{77,175,74}

\begin{document}

\begin{frontmatter}

\title{Towards Label-Efficient Geospatial Foundation Model Adaptation via Positive-Unlabeled Learning for Building Damage Mapping}

\author[inst1]{Hengwei Zhao}
\ead{hengwei.zhao@ag.tamu.edu}
\author[inst1]{Qirui Wang}
\author[inst2]{Zhuo Zheng}
\author[inst3]{Xiaoyan Lu}
\author[inst1]{Rusty Feagin}
\author[inst1]{Wenzhe Jiao\corref{cor}}
\ead{wenzhe.jiao@tamu.edu}
\affiliation[inst1]{
  organization={Department of Ecology and Conservation Biology, Texas A\&M University},
  city={College Station},
  postcode={77843}, 
  state={Texas},
  country={USA}}
\affiliation[inst2]{
  organization={Department of Computer Science, Stanford University},
  city={Stanford},
  postcode={94305}, 
  state={CA},
  country={USA}}
\affiliation[inst3]{
  organization={Department of Mechanical Engineering, National University of Singapore},
  city={Singapore},
  postcode={117575}, 
  country={Singapore}}
\cortext[cor]{Corresponding author}

\begin{abstract}
Geospatial foundation models (GFMs) are reshaping remote sensing by shifting the field from task-specific modeling toward general-purpose geospatial intelligence. Mapping the damage to buildings stands to benefit substantially from this shift, as GFMs pretrained on large-scale Earth observation data provide generalizable geospatial representations that can facilitate effective responses to sudden-onset disasters. Although parameter-efficient fine-tuning (PEFT) facilitates GFM adaptation, parameter efficiency does not necessarily translate into label efficiency. Building damage mapping remains challenging due to scarce pre-existing labeled data or limited time for annotation. When weak supervision is introduced to alleviate the labeling bottleneck, we identify the difficulty of learning sufficiently discriminative representations as a key challenge for label-efficient GFM adaptation. To address this challenge, we propose a Positive-Unlabeled GFM Adaptation (PUMA) framework that requires annotations for only a small number of damaged buildings. More specifically, PUMA integrates non-contrastive weakly supervised representation refinement to mitigate the impact of noisy supervision on representation learning from a gradient perspective, and latent-state-based label disambiguation to enable reversible supervision refinement without requiring class prior estimation. Additionally, these two components are mutually reinforcing, and their interaction can be theoretically interpreted from a generalized expectation-maximization (GEM)-inspired perspective. Extensive experiments on disaster benchmark datasets from around the world demonstrated the effectiveness and robustness of PUMA. A real-world application to the 2025 Los Angeles wildfires further highlighted its potential to accurately map damage under highly limited supervision. More broadly, these findings suggest that weakly supervised PEFT offers a label-efficient pathway from GFMs to real-world applications.
\end{abstract}



\end{frontmatter}

\section{Introduction}
Geospatial foundation models (GFMs) are reshaping remote sensing by shifting the field from task-specific model development toward general-purpose geospatial intelligence~\citep{10916803,11097335}. Advances in model architectures and large-scale pretraining enable GFMs to learn generalizable geospatial representations from vast Earth observation data, providing a transferable foundation for downstream tasks without requiring each task to be learned from scratch~\citep{10504785,brown2025alphaearth,danish2026terrafm,Feng_2026_CVPR,Herzog_2026_CVPR,FANG2026115256}. However, pretrained GFMs rarely realize their full potential without downstream adaptation~\citep{11119145,11097335,CHENG2026115694,LI2026115234,LI2026115138,TIAN2025114775}. In this context, parameter-efficient fine-tuning (PEFT) has emerged as an attractive alternative for adapting pretrained GFMs to downstream tasks~\citep{11119145}.

Among the applications that stand to benefit from this shift, mapping the damage to buildings is arguably one of the most consequential and demanding~\citep{rs17101664,ZHENG2025114979,li2026smarttransferleveragingvision,ZHENG2024114416}. Timely, accurate, and large-scale information on damaged buildings is critical for emergency decision-making, response prioritization, post-disaster loss estimation~\citep{GE2023105,ZHENG2025114979,ZHENG2021112636}, and, more broadly, disaster-resilience objectives reflected in the United Nations' Sustainable Development Goal 11~\citep{un_sdg_2015}. GFMs are particularly promising in this context because their pretrained representations capture rich geospatial features. However, parameter-efficient adaptation is not necessarily label-efficient. Most existing PEFT approaches rely on fully supervised learning, implicitly assuming the availability of reliable annotations at scale~\citep{10485462,11533530,11511776}. This assumption is difficult to satisfy in post-disaster operations, when actionable information is needed rapidly while high-quality large-scale annotations remain costly and time-consuming to acquire. This mismatch highlights a fundamental challenge in realizing the full potential of pretrained GFMs for general-purpose geospatial intelligence: achieving not only parameter-efficient adaptation but also label-efficient utilization of pretrained geospatial representations under severely limited supervision.

Integrating weakly supervised learning with PEFT offers a promising pathway toward label-efficient adaptation of GFMs. Weakly supervised learning enables models to learn from incomplete, inexact, or inaccurate supervision~\citep{10.1093/nsr/nwx106,8735810,https://doi.org/10.1049/cit2.12216}. Among various weakly supervised learning paradigms, positive-unlabeled (PU) learning~\citep{PIIROINEN2018119,ZHAO2022328,Zhao_2023_ICCV,10174705,zhao2026noisypair} is particularly well aligned with real-world disaster response workflows. In practical disaster response scenarios, obtaining comprehensive damage annotations for both damaged and undamaged buildings at scale is often prohibitively expensive and time-consuming, whereas a limited number of damaged buildings can typically be identified through field surveys, crowdsourced reports, or rapid visual interpretation. Consequently, the available training data often consist of a small set of confirmed damaged buildings (positive samples) and a large pool of unlabeled buildings with unknown damage status. This setting naturally corresponds to PU learning, which aims to train a binary classifier using only positive and unlabeled data. By eliminating the need for exhaustive negative annotations while exploiting abundant unlabeled data, PU learning provides a practical formulation for large-scale mapping of building damage under severe time and resource constraints. Accordingly, developing positive-unlabeled PEFT represents an important step toward efficiently adapting GFMs under weak supervision for operational post-disaster response.

In recent years, PU learning has been conceptualized and developed from multiple perspectives. Early studies primarily relied on two-step heuristic methods~\citep{gong2018margin}, which first identify reliable negative samples and subsequently train a binary classifier; however, their performance largely depends on the quality of the selected negatives. More recent studies have explored cost-sensitive learning~\citep{9201373}, label disambiguation~\citep{ijcai2019p590,zhao2026noisypair}, density-ratio estimation~\citep{kato2018learning}, variational optimization~\citep{NEURIPS2020_aa0d2a80,Zhao_2023_ICCV}, and other one-step formulations~\citep{xinrui2023beyond,zhou2025weaklysupervised} that jointly leverage positive samples and the entire unlabeled dataset. Among these approaches, risk-estimation-based methods~\citep{kiryo2017positive,ZHAO2022328} have emerged as a prominent paradigm due to their solid theoretical foundations and competitive empirical performance. Despite these advances, the effective integration of PU learning with PEFT for large pretrained GFMs remains largely underexplored.

\begin{figure}[!t]
\centering
\begin{subfigure}[b]{0.22\textwidth}
    \centering
    \includegraphics[width=\linewidth]{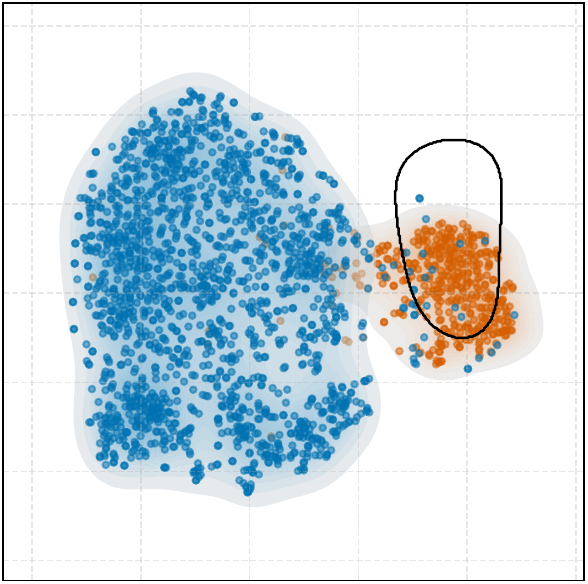}
    \caption{CE-GFM}
\end{subfigure}
\hfill
\begin{subfigure}[b]{0.22\textwidth}
    \centering
    \includegraphics[width=\linewidth]{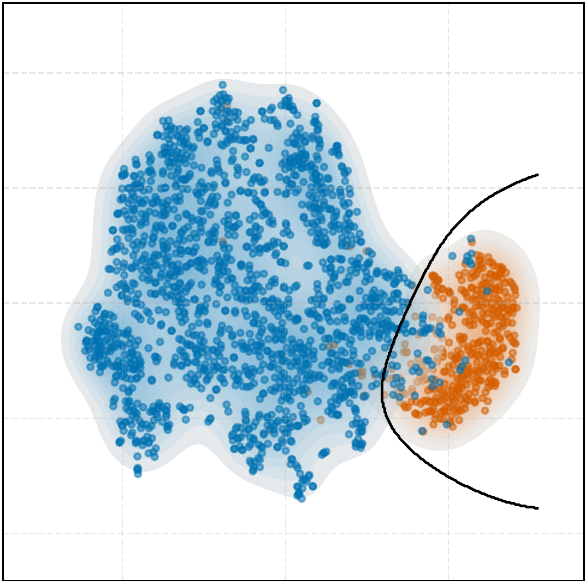}
    \caption{nnPU-GFM}
    \label{fig:nnPU_GFM}
\end{subfigure}
\hfill
\begin{subfigure}[b]{0.22\textwidth}
    \centering
    \includegraphics[width=\linewidth]{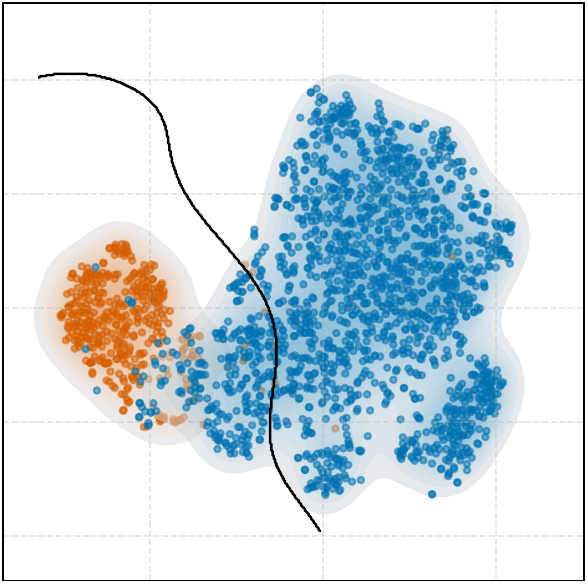}
    \caption{ROUTE}
\end{subfigure}
\hfill
\begin{subfigure}[b]{0.22\textwidth}
    \centering
    \includegraphics[width=\linewidth]{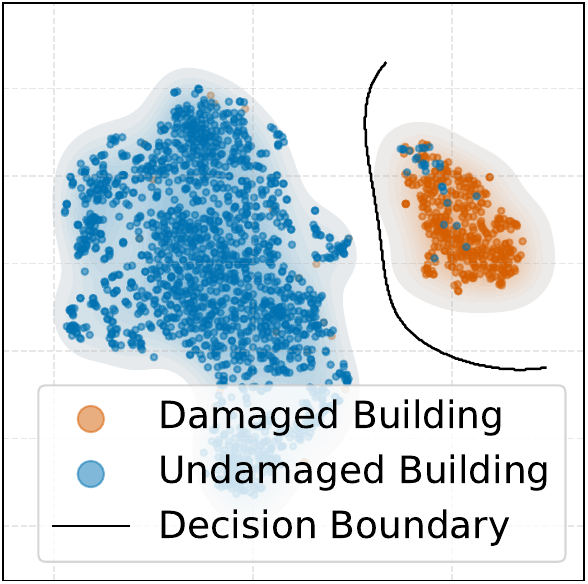}
    \caption{PUMA (ours)}
\end{subfigure}
\caption{t-SNE~\citep{tSNE} visualization of learned representations, representation distributions estimated using kernel density estimation (KDE), and decision boundaries produced by different methods on the hurricane-florence subset of the xBD dataset. VPT-deep was used as the PEFT strategy for all methods.}
\label{fig:representation_overlap}
\end{figure}

Focusing on label-efficient GFM adaptation, we identify the joint learning of discriminative representations and an accurate decision boundary as a key bottleneck in positive-unlabeled PEFT (Figure~\ref{fig:representation_overlap}). As illustrated in Figure~\ref{fig:representation_overlap}(\subref{fig:nnPU_GFM}), directly combining a PU learning method with PEFT can result in substantial overlap between the learned representations of damaged and undamaged buildings, thereby weakening class separability and limiting subsequent decision-boundary learning. This difficulty is particularly important for PEFT because only a restricted subset of parameters is updated, making effective representation adaptation under weak supervision nontrivial. Moreover, many PU learning methods~\citep{10264106,pmlr-v235-li24p,10870373} rely on the class prior ($\pi_p$), i.e., the proportion of positive samples in the underlying data distribution, which must be estimated in advance using separate methods. In the mapping of building damage, this prior corresponds to the proportion of damaged buildings among all buildings. Reliable estimation of $\pi_p$ is therefore difficult in rapidly evolving post-disaster scenarios, and estimation errors can further degrade classification performance.

To enable effective PEFT of GFMs under weak supervision, this study proposes PUMA, a positive-unlabeled geospatial foundation model adaptation framework. Inspired by recent advances in self-supervised and fully supervised representation learning~\citep{BYOL,khosla2020supervised,9578004,lu2023fmicl}, the non-contrastive weakly supervised representation refinement (NeWER) module is introduced to address the challenge of incorrect pairwise relations introduced when representation learning is incorporated into weakly supervised PEFT. Specifically, weak supervision may induce \textit{inter-class noisy pairs}, which incorrectly pull samples from different classes closer in the embedding space, and \textit{intra-class noisy pairs}, which erroneously push samples from the same class apart. To counteract these effects, a non-contrastive representation learning architecture is adapted to mitigate the impact of intra-class noisy pairs, while two representation refinement losses ($\mathcal{L}_\text{rep}^\text{sqrt}$ and $\mathcal{L}_\text{rep}^\text{log}$) are proposed to mitigate the impact of inter-class noisy pairs, with the robustness of both losses explained through gradient-based analysis. Building upon the discriminative representations learned by NeWER, the latent-state-based label disambiguation (LaTTE) module is introduced to progressively refine supervision through reversible label disambiguation, enabling PUMA to learn accurate decision boundaries without requiring class prior estimation. Notably, NeWER and LaTTE mutually reinforce each other, and their interaction admits a generalized expectation-maximization (GEM) interpretation that is consistently supported by the experimental results. Extensive experiments on disaster benchmark datasets from around the world and the 2025 Los Angeles wildfires demonstrate the effectiveness, robustness, and practical potential of PUMA for accurate large-scale damage mapping under highly limited supervision.

The remainder of this paper is organized as follows. Section~\ref{sec:area_data} introduces the study areas and datasets used in this study. Section~\ref{sec:method} presents the proposed methodology, including the overall PUMA framework and its key components. Section~\ref{sec:EM} provides a theoretical analysis of PUMA from a GEM-inspired perspective. Section~\ref{sec:experiments} reports and discusses the experimental results, and Section~\ref{sec:conclusion} concludes the paper.

\section{Study areas and datasets}\label{sec:area_data}

\begin{figure}[!t]
\centering
\includegraphics[width=0.98\linewidth]{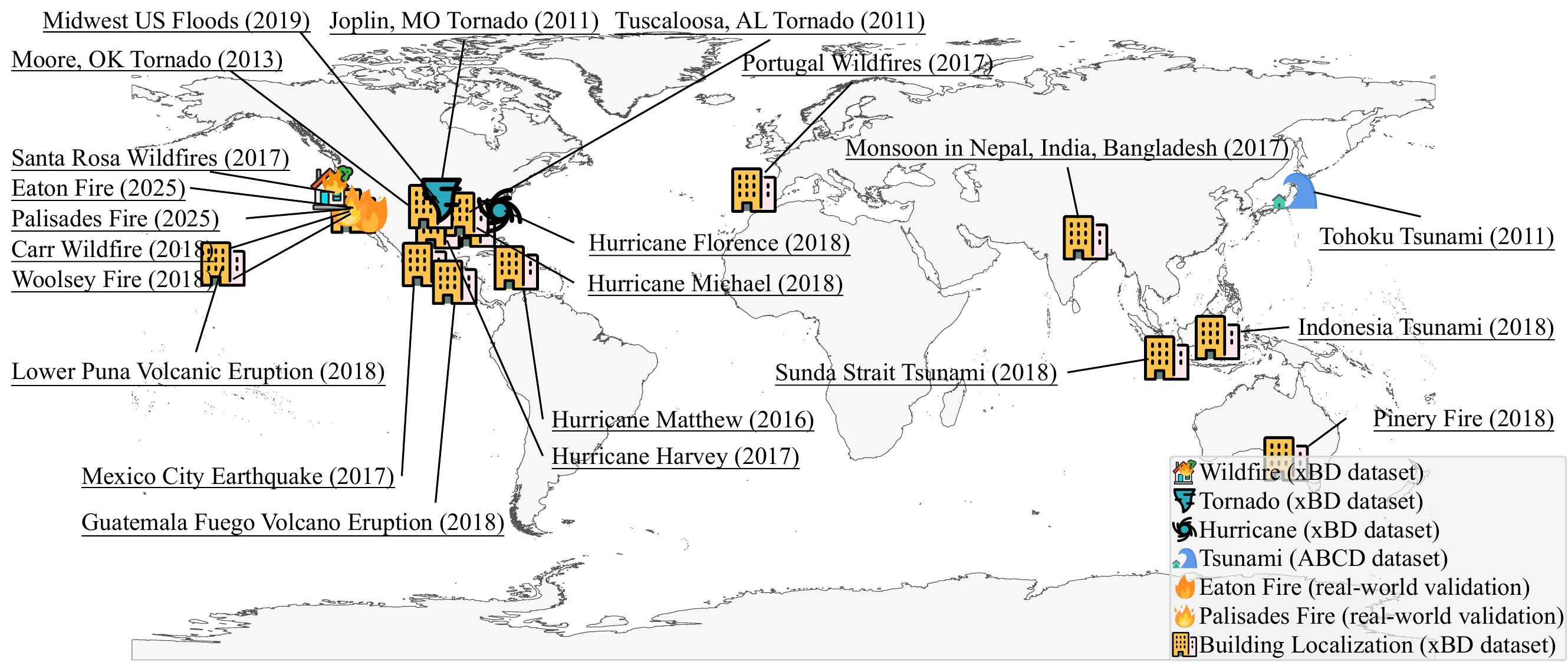}
\caption{Spatial distribution of all disaster events considered in this paper.}
\label{fig:disaster_distribution}
\end{figure}

In this study, the performance of PUMA was evaluated from two complementary perspectives: evaluation on disaster benchmark datasets from around the world and local-scale real-world validation. For benchmark evaluation, experiments were conducted on the xView2 Building Damage Assessment (xBD) dataset~\citep{Gupta_2019_CVPR_Workshops} and the AIST Building Change Detection (ABCD) dataset~\citep{7986759}. This evaluation included quantitative comparisons with state-of-the-art methods and comprehensive ablation studies to assess the contribution of each module of the proposed framework. To further evaluate its practical applicability, a local-scale validation was performed using data from the 2025 Eaton Fire and the 2025 Palisades Fire, two major events of the 2025 Los Angeles wildfires, which rank among the most destructive wildfires in California's history. These experiments further demonstrate the effectiveness of PUMA for operational post-disaster building damage mapping under real-world conditions. The spatial distribution of all disaster events considered in this study is shown in Figure~\ref{fig:disaster_distribution}.

\subsection{Benchmark study areas and datasets}

\subsubsection{xView2 building damage assessment dataset}
As shown in Figure~\ref{fig:disaster_distribution}, the xBD dataset is a global-scale benchmark for building damage assessment, spanning four continents: Asia, Europe, North America, and Oceania. It contains 11,034 pairs of high-resolution optical satellite images, covering 45,361.79 km$^2$ and encompassing 850,736 building footprints. Each image pair consists of a pre-disaster image and its corresponding post-disaster image. The imagery was acquired from multiple satellite platforms, including GeoEye, WorldView-2, and WorldView-3, resulting in varying ground sample distances (GSDs), all finer than 0.8 m. Building damage in the xBD dataset is annotated according to the Joint Damage Scale, which was developed based on the HAZUS natural hazard analysis framework~\citep{vickery2006hazus}, the Kelman scale~\citep{kelman2003physical}, and the EMS-98 scale~\citep{grunthal1998european}. To formulate building damage assessment as a binary classification task and maintain label consistency with the ABCD dataset, following the protocol in~\cite{xia2021}, the major damage and destroyed categories were merged into the damaged class, whereas the no damage and minor damage categories were combined into the undamaged class. Three disaster events, namely hurricane-florence (hurricane), joplin-tornado (tornado), and santa-rosa-wildfire (wildfire), were selected to evaluate the effectiveness of the proposed PUMA framework. The pre-disaster imagery from the remaining xBD disaster events was used to train the building footprint extraction model (described in detail in Section~\ref{sec:building_localization}). Following the official xBD data split, 10\% of the image pairs from each selected event were reserved for model evaluation. For GFM adaptation, only 300 damaged buildings together with 5,000 unlabeled buildings were used for PEFT. Detailed information about the xBD dataset is summarized in Table~\ref{tab:bd_summary}.

\begin{table}[!b]
\centering
\caption{Summary statistics of the xBD dataset.}
\label{tab:bd_summary}
\resizebox{\linewidth}{!}{%
\begin{tblr}{
  row{odd} = {c},
  row{2} = {font=\bfseries},
  row{4} = {c},
  row{6} = {font=\bfseries},
  row{8} = {c},
  row{10} = {c},
  row{12} = {c},
  row{14} = {c},
  row{16} = {c},
  row{18} = {c},
  row{20} = {c},
  row{22} = {c},
  cell{2}{1} = {c=4}{},
  cell{6}{1} = {c=4}{},
  hline{1-3,6-7,23} = {-}{},
}
Name                                                                                & Disaster event                      & Event date            & Number of images \\
Used for building damage classification and building localization evaluation &                                     &                       &                       \\
hurricane-florence                                                                  & Hurricane Florence                  & Sep 10-19, 2018       & 1092                  \\
joplin-tornado                                                                      & Joplin, MO Tornado                  & May 22, 2011          & 298                   \\
santa-rosa-wildfire                                                                 & Santa Rosa Wildfires                & Oct 8-31, 2017        & 754                   \\
Used only for training the building localization model                              &                                     &                       &                       \\
guatemala-volcano                                                                   & Guatemala Fuego Volcano Eruption    & Jun 3, 2018           & 56                    \\
hurricane-harvey                                                                    & Hurricane Harvey                    & Aug 17 - Sep 2, 2017  & 1044                  \\
hurricane-matthew                                                                   & Hurricane Matthew                   & Sep 28 - Oct 10, 2016 & 810                   \\
hurricane-michael                                                                   & Hurricane Michael                   & Oct 7-16, 2018        & 1100                  \\
mexico-earthquake                                                                   & Mexico City Earthquake              & Sep 19, 2017          & 386                   \\
midwest-flooding                                                                    & Midwest US Floods                   & Jan 3 - May 31, 2019  & 890                   \\
palu-tsunami                                                                        & Indonesia Tsunami                   & Sep 18, 2018          & 392                   \\
socal-fire                                                                          & Carr Wildfire                       & Jul 23 - Aug 30, 2018 & 2806                  \\
lower-tuna-volcano                                                                  & Lower Puna Volcanic Eruption        & May 23 - Aug 14, 2018 & 582                   \\
moore-tornado                                                                       & Moore, OK Tornado                   & May 20, 2013          & 454                   \\
nepal-flooding                                                                      & Monsoon in Nepal, India, Bangladesh & Jul - Sep, 2017       & 1238                  \\
pinery-bushfire                                                                     & Pinery Fire                         & Nov 25 - Dec 2, 2018  & 3690                  \\
portugal-wildfire                                                                   & Portugal Wildfires                  & Jun 17-24, 2017       & 3738                  \\
sunda-tsunami                                                                       & Sunda Strait Tsunami                & Dec 22, 2018          & 296                   \\
tuscaloosa-tornado                                                                  & Tuscaloosa, AL Tornado              & Apr 27, 2011          & 686                   \\
woolsey-fire                                                                        & Woolsey Fire                        & Nov 9-28, 2018        & 1756                  
\end{tblr}
}
\end{table}

\subsubsection{AIST building change detection dataset}
The ABCD dataset was developed to determine whether buildings were destroyed by the 2011 Tohoku tsunami in Japan. It contains 4,221 damaged (washed-away) and 4,221 undamaged buildings, represented by pre-/post-event image pairs covering approximately 66 km$^2$. All images have a GSD of 0.4 m. The pre-tsunami imagery was acquired in August 2000, whereas the post-tsunami imagery was acquired within one month after March 11, 2011. Each pre-/post-event image pair is centered on a single building, with the damage status of the central building (i.e., whether it was washed away) serving as the corresponding label. Forty percent of the data were held out for evaluation. For GFM adaptation, only 300 damaged buildings and 4,000 unlabeled buildings were used for PEFT.

\subsection{Real-world case study areas and data}
The January 2025 Southern California wildfires rank among the most destructive wildfire disasters in recent U.S. history, causing severe impacts across Los Angeles and surrounding areas. Driven by prolonged dry conditions, low humidity, and strong Santa Ana winds, the fires spread rapidly and resulted in widespread damage to buildings and critical infrastructure, localized power outages, and substantial loss of life. Among the major wildfire events, the Eaton Fire and the Palisades Fire were two of the most destructive and were therefore selected as representative case studies for evaluating the practical applicability of PUMA.

Satellite imagery for the Eaton Fire and the Palisades Fire was acquired from multiple sources. For the Eaton Fire, the pre-disaster imagery was obtained from Bing Maps, whereas the post-disaster imagery was acquired by the WorldView-3 (WV03) satellite on January 14, 2025, with an original GSD of 0.32 m. Both pre- and post-disaster images were resampled to a uniform GSD of 0.3 m, resulting in image dimensions of $16,932\times17,075$ pixels. For the Palisades Fire, the pre-disaster imagery was also obtained from Bing Maps, while the post-disaster imagery was acquired by the WorldView-2 (WV02) satellite on January 14, 2025, with an original GSD of 0.57 m. Similarly, both images were resampled to a GSD of 0.3 m, yielding final dimensions of $17,314\times17,428$ pixels.

The initial ground-truth annotations were derived from Microsoft's building damage assessment products (Eaton Fire: \url{https://data.humdata.org/dataset/eaton-fire-altadena-damage-assessment-from-1-10}, Palisades Fire: \url{https://data.humdata.org/dataset/palisades-fire-building-damage-assessment}). The annotations for the two study areas were subsequently refined by two independent expert teams. Each team's revisions were cross-checked by the other team through an iterative review process. Buildings with inconsistent labels were further verified using high-resolution imagery, while buildings whose damage status remained ambiguous after review were excluded from the final ground-truth set. This process was repeated until consensus was reached between the two teams, resulting in the final reference annotations. For GFM adaptation, 300 damaged buildings and 4,000 unlabeled buildings were used for PEFT. In addition, a more label-constrained setting comprising only 50 damaged buildings and 500 unlabeled buildings was considered to further assess the effectiveness of PUMA under limited supervision.

\section{Methodology}\label{sec:method}

\subsection{Problem reformulation for building damage mapping}

\begin{figure}[!t]
\centering
\includegraphics[width=0.98\linewidth]{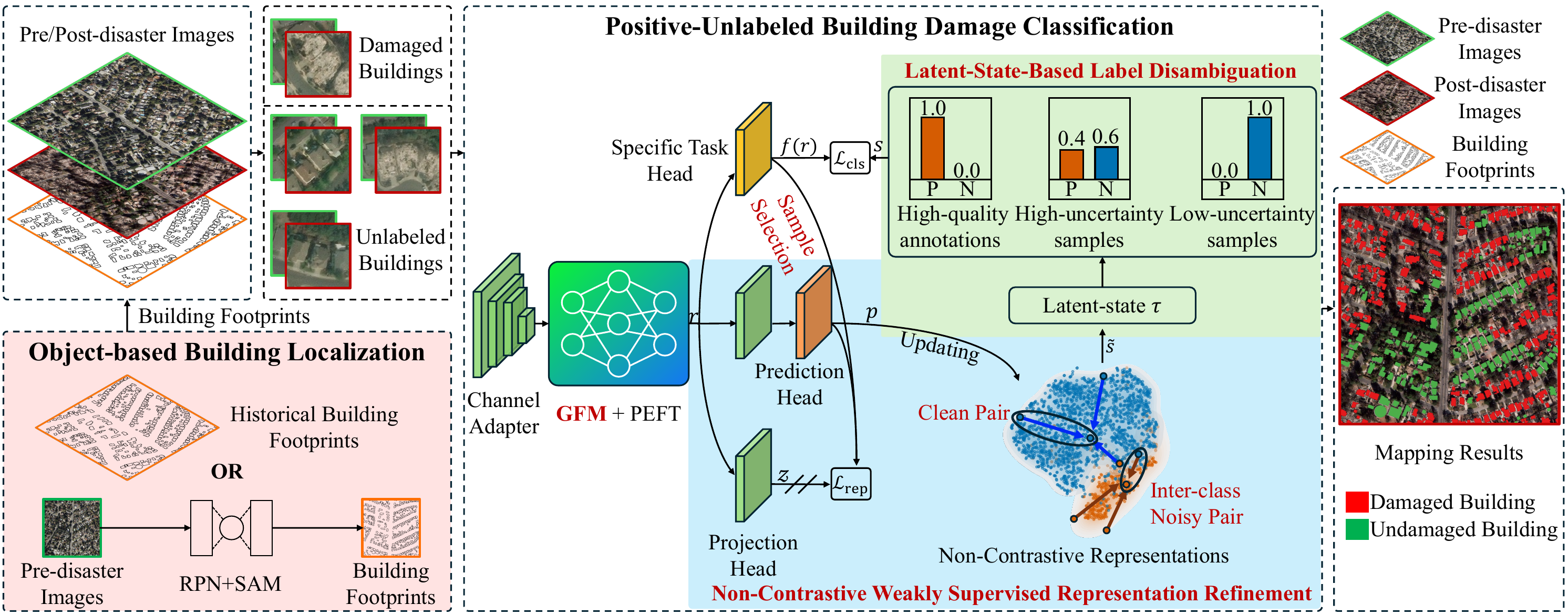}
\caption{Overview of the proposed positive-unlabeled GFM adaptation framework.}
\label{fig:framework}
\end{figure}

Since each building is associated with a unique damage state, the mapping of building damage is inherently an object-centric problem rather than a pixel-wise classification task. Accordingly, it can be reformulated as an object-based mapping problem, which comprises two complementary subtasks: object-based building localization and building damage classification. The former identifies individual building instances, whereas the latter determines the damage state of each localized building. In real-world disaster response, however, obtaining comprehensive building-level damage annotations is often prohibitively expensive and time-consuming. Therefore, building damage classification is naturally formulated as a PU learning problem. The inputs and outputs of the two subtasks are illustrated in Figure~\ref{fig:framework}.

\subsubsection{Object-based building localization}\label{sec:building_localization}
Building localization aims to delineate individual building footprints at the instance level, thereby providing the object units required for subsequent object-based building damage classification. To enhance applicability across diverse scenarios, a hybrid localization strategy is adopted. Historical building footprints are directly utilized when reliable and up-to-date footprint data are available. Otherwise, building instances are automatically delineated from pre-disaster imagery using an integrated Region Proposal Network (RPN)~\citep{8237584} and Segment Anything Model (SAM)~\citep{10378323}, following the strategy in~\cite{WANG2024707}. Benefiting from the regular geometric characteristics of buildings, the increasing availability of global-scale building footprint datasets~\citep{sirko2021continentalscalebuildingdetectionhigh,microsoft2023globalbuildings,zhu2026globalbuildingmap}, and the strong representation capability of foundation models~\citep{10378323,ICLR2025_45c1f6a8}, RPN+SAM generalizes effectively across different geographic regions. Experimental results presented in this study further demonstrate that accurate building localization can be achieved in disaster-affected areas without requiring post-disaster localization annotations, facilitating the deployment of the proposed framework for large-scale disaster response.

\subsubsection{Positive-unlabeled building damage classification}
Let $\bm{x}_i \in \mathbb{R}^d$ denote a building-centered image patch cropped from channel-wise stacked pre-/post-event satellite imagery, and let $y_i$ denote its damage state, where $y_i=0$ and $y_i=1$ represent damaged and undamaged buildings, respectively. Given the training dataset $\mathcal{D}=\mathcal{D}_{p} \cup \mathcal{D}_{u}$, where $\mathcal{D}_{p}=\{(\bm{x}_i,y_i=0)\}_{i=1}^{n_p}$ and $\mathcal{D}_{u}=\{\bm{x}_i\}_{i=1}^{n_u}$, $n_p$ and $n_u$ denote the number of damaged and unlabeled buildings, respectively. The objective of positive-unlabeled building damage classification is to learn a binary classifier $f(\bm{r}_{i}):\mathbb{R}^d \rightarrow [0,1]^2$ from representations $\bm{r}_{i}=E(\bm{x}_{i})$ extracted by a GFM-based encoder $E$. The encoder comprises a channel adapter and a pretrained GFM equipped with PEFT modules. The channel adapter aligns the channel dimensionality of the stacked pre- and post-disaster imagery with the input specification of the pretrained GFM through a two-dimensional convolutional layer followed by batch normalization (BN). During fine-tuning, the parameters of the pretrained GFM remain frozen. Although GFMs hold considerable promise for building damage classification, adapting them via PEFT under unreliable supervision remains fundamentally challenging, as it requires jointly learning discriminative representations and establishing an accurate decision boundary without reliable supervision (Figure~\ref{fig:representation_overlap}).

Let $\bm{s}_{i} \in [0,1]^2$ denote the pseudo-label of $\bm{x}_{i}$, which is expected to become increasingly accurate as GFM fine-tuning progresses. Given $\bm{s}_{i}$, the classification objective can be reformulated as:
\begin{equation}
  \mathcal{L}_\text{cls}(f(\bm{r}_{i}),\bm{s}_{i})=-\bm{s}_{i}^{\top}\log f(\bm{r}_{i}).
  \label{eq:loss_cls}
\end{equation}

The reformulation in Eq.~(\ref{eq:loss_cls}) provides a simple and flexible means of performing positive-unlabeled building damage classification without requiring explicit estimation of the class prior. However, its practical effectiveness is constrained by the mutual dependence between representation learning and pseudo-label refinement during PEFT. Specifically, insufficiently discriminative representations lead to unreliable pseudo-labels, while erroneous pseudo-labels, in turn, impede the learning of discriminative representations. This dilemma is alleviated in PUMA through the mutually reinforcing interaction between NeWER and LaTTE, as described in the following section.

\subsection{PUMA: the positive-unlabeled GFM adaptation framework}
PUMA is designed to bridge the gap between weak supervision and PEFT of GFMs by facilitating discriminative representation learning under unreliable supervision. It consists of two mutually reinforcing modules: NeWER and LaTTE. Specifically, NeWER learns discriminative representations that facilitate label disambiguation, whereas LaTTE provides increasingly reliable supervision that, in turn, further enhances representation learning. This mutually reinforcing interaction can be theoretically interpreted from a GEM-inspired perspective. Importantly, PUMA does not require explicit estimation of the class prior, enhancing its practicality for real-world disaster response. An overview of the proposed PUMA framework is presented in Figure~\ref{fig:framework}.

\subsubsection{Non-contrastive weakly supervised representation refinement}

Incorrect pairwise relationships pose two major challenges to representation learning in weakly supervised PEFT: inter-class noisy pairs and intra-class noisy pairs. Inter-class noisy pairs erroneously pull samples from different classes closer in the embedding space, whereas intra-class noisy pairs incorrectly push samples from the same class apart.

\noindent \textbf{Non-contrastive representation learning architecture.} In contrast to contrastive representation learning frameworks~\citep{9157636,pmlr-v119-chen20j,khosla2020supervised,lu2023fmicl}, which rely on both sample pairs from the same class and sample pairs from different classes, non-contrastive representation learning frameworks~\citep{BYOL,9578004,9709990,oquab2024dinov} exclusively exploit sample pairs from the same class. Consequently, they inherently avoid the adverse effects of intra-class noisy pairs, making them particularly well suited for representation learning in weakly supervised PEFT.

For simplicity and without loss of generality, the NeWER module adopts a SimSiam-like architecture~\citep{9578004}. Given a building patch $\bm{x}_{i}$, two randomly augmented views, $\bm{x'}_{i}$ and $\bm{x''}_{i}$, are encoded by a shared encoder, followed by a projection head $h$. A prediction head $g$ is further applied to the projected embedding of one view. Let $\bm{z}_{i}=h(\bm{r}_{i})$ denote the projected embedding and $\bm{p}_{i}=g(\bm{z}_{i})$ denote the corresponding prediction. The self-supervised representation learning objective minimizes the distance between the two views without requiring label supervision:
\begin{equation}
  \mathcal{L}_\text{rep}^\text{self}(\bm{p}_{i},\sg(\bm{z}_i))=\norm{\bm{\tilde{p}}_{i}-\bm{\tilde{z}}_{i}}^2=2(1-\bm{\tilde{p}}_{i}^{\top}\bm{\tilde{z}}_{i}),
  \label{eq:loss_self_rep}
\end{equation}
where $\bm{\tilde{p}}_{i}=\bm{p}_{i}/\norm{\bm{p}_{i}}$ and $\bm{\tilde{z}}_{i}=\bm{z}_{i}/\norm{\bm{z}_{i}}$ denote the $\ell_{2}$-normalized prediction and projection, respectively. Here, $\sg(\cdot)$ denotes the stop-gradient operation, and $\norm{\cdot}$ denotes the $\ell_{2}$ norm. Under full supervision, label information can be incorporated into representation learning:
\begin{equation}
  \mathcal{L}_\text{rep}^\text{sup}(\bm{p}_{i},\sg(\bm{z}_j))=2(1-\bm{\tilde{p}}_{i}^{\top}\bm{\tilde{z}}_{j})\mathbb{I}\{y_i = y_j\},
  \label{eq:loss_full_rep}
\end{equation}
where $\mathbb{I}\{\cdot\}$ is the indicator function. In practice, due to the lack of full supervision, pseudo-labels generated by the classifier can be used to make Eq.~(\ref{eq:loss_full_rep}) computable, but this inevitably introduces inter-class noisy pairs.

\noindent \textbf{Representation refinement losses.} Although non-contrastive representation learning eliminates the adverse effects of intra-class noisy pairs, inter-class noisy pairs remain unavoidable when samples from different classes are erroneously assigned the same label. Such noisy pairs may produce disproportionately large gradient magnitudes and thereby bias representation learning toward incorrect supervision during PEFT. To alleviate this issue, two representation refinement losses are introduced to suppress the influence of inter-class noisy pairs.

The gradient magnitude of $\mathcal{L}_\text{rep}^\text{sup}(\bm{p}_{i},\sg(\bm{z}_{j}))$ with respect to $\bm{p}_i$ can be expressed as:
\begin{equation}
  \norm{\frac{\partial \mathcal{L}_\text{rep}^\text{sup}(\bm{p}_{i},\sg(\bm{z}_{j}))}{\partial \bm{p}_{i}}} = \frac{2}{\norm{\bm{p}_{i}}} \sqrt{1 - (\tilde{\bm{p}}_{i}^\top \tilde{\bm{z}}_{j})^2}.
\label{eq:magnitudes_full_rep}
\end{equation}
Given a clean pair $(\bm{x}_{i},\bm{x}_{j})$ and an inter-class noisy pair $(\bm{x}_{i},\bm{x}_{m})$, their embeddings may satisfy $\bm{\tilde{p}}_{i}^{\top}\bm{\tilde{z}}_{j} > \bm{\tilde{p}}_{i}^{\top}\bm{\tilde{z}}_{m}$. From a gradient perspective, some inter-class noisy pairs may induce larger gradient magnitudes than clean pairs, biasing the optimization toward fitting noisy supervision. For example, when $\bm{\tilde{p}}_{i}^{\top}\bm{\tilde{z}}_{j} \to 1$ and $\bm{\tilde{p}}_{i}^{\top}\bm{\tilde{z}}_{m} \approx 0$, the corresponding gradient magnitudes satisfy
\begin{equation}
\begin{aligned}
    \norm{\frac{\partial \mathcal{L}_\text{rep}^\text{sup}(\bm{p}_{i},\sg(\bm{z}_j))}{\partial \bm{p}_i}}&={\frac{2}{\norm{\bm{p}_{i}}}}{\sqrt{1-(\tilde{\bm{p}}_{i}^{\top}{\tilde{\bm{z}}}_{j})^2}}\\ &< {\frac{2}{\norm{\bm{p}_{i}}}}{\sqrt{1-(\tilde{\bm{p}}_{i}^{\top}{\tilde{\bm{z}}}_{m})^2}}= \norm{\frac{\partial \mathcal{L}_\text{rep}^\text{sup}(\bm{p}_{i},\sg(\bm{z}_m))}{\partial \bm{p}_i}}.
\end{aligned}
\end{equation}

To mitigate this problem, we build on and extend our previous work~\citep{zhao2026noisypair} by introducing two representation refinement losses, $\mathcal{L}_\text{rep}^\text{sqrt}$ and $\mathcal{L}_\text{rep}^\text{log}$. Specifically, the square-root-based loss $\mathcal{L}_\text{rep}^\text{sqrt}(\bm{p}_{i},\sg(\bm{z}_j))$ is defined as
\begin{equation}
  \mathcal{L}_\text{rep}^\text{sqrt}(\bm{p}_{i},\sg(\bm{z}_j))=2\sqrt{k(1-\bm{\tilde{p}}_{i}^{\top}\bm{\tilde{z}}_{j})}\mathbb{I}\{y_i = y_j\},
  \label{eq:loss_weak_rep_sqrt}
\end{equation}
where $k>0$ is a hyperparameter controlling how tightly the representations are clustered around the mean direction, as described in Section~\ref{sec:EM}. The logarithm-based loss $\mathcal{L}_\text{rep}^\text{log}(\bm{p}_{i},\sg(\bm{z}_j))$ is formulated as
\begin{equation}
  \mathcal{L}_\text{rep}^\text{log}(\bm{p}_{i},\sg(\bm{z}_j))=2\log{(1-\bm{\tilde{p}}_{i}^{\top}\bm{\tilde{z}}_{j})}\mathbb{I}\{y_i = y_j\}.
  \label{eq:loss_weak_rep_log}
\end{equation}
$\mathcal{L}_\text{rep}^\text{log}(\bm{p}_{i},\sg(\bm{z}_j))$ and $\mathcal{L}_\text{rep}^\text{sqrt}(\bm{p}_{i},\sg(\bm{z}_{j}))$ also aim to align embeddings of same-class data, while the gradient magnitudes of clean pairs are larger than those of noisy pairs:
\begin{equation}
  \begin{aligned}
    \norm{\frac{\partial \mathcal{L}_\text{rep}^\text{sqrt}(\bm{p}_i,\sg(\bm{z}_j))}{\partial \bm{p}_i}} &= \frac{\sqrt{k}}{\norm{\bm{p}_i}}\sqrt{1 + \tilde{\bm{p}}_i^\top \tilde{\bm{z}}_j} \\
    &> \frac{\sqrt{k}}{\norm{\bm{p}_i}}\sqrt{1 + \tilde{\bm{p}}_i^\top \tilde{\bm{z}}_m} = \norm{\frac{\partial \mathcal{L}_\text{rep}^\text{sqrt}(\bm{p}_i,\sg(\bm{z}_m))}{\partial \bm{p}_i}},
  \end{aligned}
\end{equation}
\begin{equation}
  \begin{aligned}
    \norm{\frac{\partial \mathcal{L}_\text{rep}^\text{log}(\bm{p}_i,\sg(\bm{z}_j))}{\partial \bm{p}_i}} &= \frac{2}{\norm{\bm{p}_i}}\sqrt{\frac{1 + \tilde{\bm{p}}_i^\top \tilde{\bm{z}}_j}{1 - \tilde{\bm{p}}_i^\top \tilde{\bm{z}}_j}} \\
    &> \frac{2}{\norm{\bm{p}_i}}\sqrt{\frac{1 + \tilde{\bm{p}}_i^\top \tilde{\bm{z}}_m}{1 - \tilde{\bm{p}}_i^\top \tilde{\bm{z}}_m}} = \norm{\frac{\partial \mathcal{L}_\text{rep}^\text{log}(\bm{p}_i,\sg(\bm{z}_m))}{\partial \bm{p}_i}}.
  \end{aligned}
\end{equation}
Experimental results demonstrate that the PUMA framework achieves promising performance with either $\mathcal{L}_\text{rep}^\text{sqrt}(\bm{p}_{i},\sg(\bm{z}_j))$ or $\mathcal{L}_\text{rep}^\text{log}(\bm{p}_{i},\sg(\bm{z}_{j}))$ in Section~\ref{sec:experiments}. Detailed derivations of the above results are provided in~\ref{apx:gradient}.

\subsubsection{Latent-state-based label disambiguation}

Building on the increasingly discriminative representations learned by the NeWER module, the LaTTE module is designed to progressively refine the supervision used for GFM fine-tuning. Specifically, LaTTE integrates class prototype-based label disambiguation with latent-state-based pseudo-label refinement, enabling the conservative introduction of negative supervision without undermining the effectiveness of prototype-based label disambiguation.

\noindent \textbf{Label disambiguation via class prototypes.} The class prototype $\tilde{\bm{\nu}}_{c}$, which serves as the representative embedding of class $c$ in the embedding space, is updated using an exponential moving average (EMA):
\begin{equation}
  \tilde{\bm{\nu}}_{c} = \frac{\alpha\tilde{\bm{\nu}}_{c}+(1-\alpha)\tilde{\bm{p}}_{i}}{\norm{\alpha\tilde{\bm{\nu}}_{c}+(1-\alpha)\tilde{\bm{p}}_{i}}},
\end{equation}
where $\tilde{\bm{p}}_{i}$ denotes the $\ell_2$-normalized prediction of sample $\bm{x}_{i}$, and $\alpha$ is the momentum coefficient. The prototype $\tilde{\bm{\nu}}_{c}$ is updated whenever $\bm{x}_{i}$ is assigned to class $c$.

Owing to the discriminative representations learned by NeWER, the class prototypes can be leveraged for label disambiguation, thereby producing more reliable supervision:
\begin{equation}
    \tilde{\bm{s}}_{i} = \beta\tilde{\bm{s}}_{i}+(1-\beta)\bm{\rho}_{i}, \quad \bm{\rho}_{i} = \OneHot(\arg\max_{c}\tilde{\bm{p}}_{i}^{\top}\tilde{\bm{\nu}}_{c}),
\end{equation}
where $\OneHot(\cdot)$ converts the predicted class index into a one-hot vector. However, the effectiveness of prototype-based label disambiguation depends on the availability of the class prior $\pi_p$~\citep{10870373}. In the absence of $\pi_p$, a lack of sufficiently reliable negative supervision can bias the GFM toward classifying undamaged buildings as damaged (as further demonstrated in the ablation study), ultimately leading to a trivial solution.

\noindent \textbf{Latent-state-based pseudo-label refinement.} To eliminate the dependence on $\pi_p$, reliable negative supervision is explicitly introduced during GFM fine-tuning. Specifically, the prediction confidence of the GFM is modeled as a latent state ($\tau$), which reflects the reliability of its current predictions. An unlabeled sample is regarded as an undamaged sample when its predicted probability of belonging to the undamaged class is greater than or equal to $\tau$. Accordingly, the pseudo-label of an unlabeled sample $\bm{x}_{i}$ is defined as
\begin{equation}
  \bm{s}_{i}=
  \begin{cases}
    [1,0]^\top,
    & \bm{x}_{i}\in\mathcal{D}_{p},\\
    \tilde{\bm{s}}_{i},
    & \bm{x}_{i}\in\mathcal{D}_{u} \quad \text{and} \quad P_{\mathrm{GFM}}(y_{i}=1\mid\bm{x}_{i})<\tau,\\
    [0,1]^\top,
    & \bm{x}_{i}\in\mathcal{D}_{u} \quad \text{and} \quad P_{\mathrm{GFM}}(y_{i}=1\mid\bm{x}_{i}) \ge \tau.
\end{cases}
\end{equation}
To avoid manually specifying $\tau$, self-adaptive threshold (SAT)~\citep{wang2023freematch} is adopted to automatically estimate the confidence threshold from the prediction statistics of the GFM during fine-tuning. Specifically, $\tau$ starts from a relatively low value, allowing more unlabeled samples to contribute undamaged supervision during the early stage of fine-tuning, and gradually increases as training progresses to filter out less reliable predictions. The pseudo-label of each unlabeled building is initialized as $[0,1]^\top$.

The proposed latent-state-based pseudo-label refinement provides a more conservative mechanism for introducing reliable negative supervision. In particular, if an unlabeled building is incorrectly predicted as undamaged with high confidence, but this prediction is later identified as mislabeled, its pseudo-label reverts to $\tilde{\bm{s}}_{i}$ and is subsequently refined through prototype-based label disambiguation, rather than being replaced by the hard undamaged label $[0,1]^\top$. For example, suppose that at the $t$-th update, an unlabeled sample $\bm{x}_i$ is assigned the hard pseudo-label $[0,1]^\top$ because $P_{\mathrm{GFM}}(y_{i}=1\mid\bm{x}_{i}) \ge \tau$. At the subsequent $(t+1)$-th update, however, the GFM becomes less confident in its prediction for $\bm{x}_i$, such that $P_{\mathrm{GFM}}(y_{i}=1\mid\bm{x}_{i})<\tau$. Rather than continuing to update the pseudo-label from the previously assigned hard label $[0,1]^\top$, PUMA reverts to $\tilde{\bm{s}}_{i}$ as the starting point for further pseudo-label refinement.

\subsubsection{Overall objective}
The overall optimization objective of PUMA for GFM fine-tuning is formulated as
\begin{equation}
  \mathcal{L}_\text{all}=\frac{1}{\lvert \mathcal{B}_{p} \rvert}\sum_{\bm{x}_{i} \in \mathcal{B}_{p}}\mathcal{L}_\text{cls}+\frac{1}{\lvert \mathcal{B}_{u} \rvert}\sum_{\bm{x}_{i} \in \mathcal{B}_{u}}\mathcal{L}_\text{cls}+w_{rep}\frac{1}{\lvert \mathcal{B} \rvert}\sum_{\bm{x}_{i} \in \mathcal{B}}\frac{1}{\lvert \mathcal{Z}_{i} \rvert}\sum_{\bm{x}_{j} \in \mathcal{Z}_{i}}\mathcal{L}_\text{rep},
\end{equation}
where $\mathcal{B}_{p}$ and $\mathcal{B}_{u}$ denote the damaged and unlabeled samples in mini-batch $\mathcal{B}$, respectively, and $\mathcal{Z}_{i}$ denotes the set of samples assigned to the same class as $\bm{x}_{i}$. $\mathcal{L}_\text{rep}$ can be instantiated as either $\mathcal{L}_\text{rep}^\text{sqrt}$ or $\mathcal{L}_\text{rep}^\text{log}$. The theoretical analysis in Section~\ref{sec:EM} demonstrates that, when $\mathcal{L}_\text{rep}^\text{sqrt}$ is employed, the PUMA framework can be interpreted as a GEM-inspired alternating optimization framework. Therefore, unless otherwise specified, $\mathcal{L}_\text{rep}^\text{sqrt}$ is adopted as the representation refinement loss throughout the remainder of this paper. An entropy regularization term is incorporated to stabilize training.

\section{Why PUMA improves positive-unlabeled GFM fine-tuning}\label{sec:EM}
To establish a theoretical foundation for positive-unlabeled GFM adaptation, the alternating optimization procedure of PUMA is interpreted within a GEM-inspired framework when $\mathcal{L}_\text{rep}^\text{sqrt}$ is employed as the representation refinement loss. By parameterizing the latent posterior using the classifier predictions, the variational lower bound on the unlabeled-data log-likelihood can be connected to the representation refinement loss under mild assumptions. In contrast to existing analyses~\citep{10870373,zhao2026noisypair}, the following derivation establishes this connection without assuming a uniform class prior. Moreover, the theoretical findings are empirically supported in the context of GFM fine-tuning. Since the labeled damaged samples have reliable annotations, the following analysis focuses on the unlabeled data. Detailed derivations are provided in~\ref{apx:EM}.

\noindent \textbf{Latent posterior inference: E-Step.} Let $\bm{\theta}$ denote the trainable parameters of the GFM and its adaptation components. By introducing a variational distribution $\mathcal{Q}(y_{i})$ over the latent class label $y_{i}$, the unlabeled-data log-likelihood can be lower-bounded as follows:
\begin{equation}
  \begin{aligned}
    \mathcal{J}(\bm{\theta})
    &=\sum_{\bm{x}_{i} \in \mathcal{D}_{u}}\log \sum_{y_{i} \in \mathcal{Y}}p(\bm{x}_{i}, y_{i};\bm{\theta})\\
    &=\sum_{\bm{x}_{i} \in \mathcal{D}_{u}}\log \sum_{y_{i} \in \mathcal{Y}}\mathcal{Q}(y_{i})\frac{p(\bm{x}_{i}, y_{i};\bm{\theta})}{\mathcal{Q}(y_{i})}\\
    &\ge \sum_{\bm{x}_{i} \in \mathcal{D}_{u}} \sum_{y_{i} \in \mathcal{Y}}\mathcal{Q}(y_{i})\log \frac{p(\bm{x}_{i}, y_{i};\bm{\theta})}{\mathcal{Q}(y_{i})}\\
    &\coloneqq \mathrm{ELBO}(\bm{\theta},\mathcal{Q}).
  \end{aligned}
  \label{eq:log-likelihood}
\end{equation}

The gap between the log-likelihood and the evidence lower bound (ELBO) is given by
\begin{equation}
  \mathcal{J}(\bm{\theta})-\mathrm{ELBO}(\bm{\theta},\mathcal{Q})=\sum_{\bm{x}_{i} \in \mathcal{D}_{u}}\mathrm{KL}(\mathcal{Q}(y_{i}) \parallel p(y_{i} \mid \bm{x}_{i};\bm{\theta})).
\end{equation}
Therefore, the ELBO becomes tight when the variational distribution coincides with the posterior distribution:
\begin{equation}
    \mathcal{Q}(y_{i})=p(y_{i} \mid \bm{x}_{i};\bm{\theta}).
\end{equation}
For simplicity, the E-step is approximated using a hard-assignment strategy: $\mathcal{Q}(y_i)=\mathbb{I}\{\tilde{y}_{i}=y_{i}\}$, where $\tilde{y}_{i}=\mathop{\arg\max}f(\bm{r}_{i})$ denotes the predicted class label of $\bm{x}_{i}$.

\noindent \textbf{Log-likelihood maximization: M-step.} In the M-step, under mild assumptions, minimizing the proposed representation refinement loss can be interpreted as maximizing a lower bound on the unlabeled-data log-likelihood. For analytical convenience, all unlabeled samples are considered at each iteration and partitioned into class-specific subsets $\mathcal{U}_{c}=\{ \bm{x}_{i} \in \mathcal{D}_{u} \mid \tilde{y}_{i}=c \}$:
\begin{equation}
  \tilde{\mathcal{R}}(\bm{x})=\frac{1}{n_{u}}\sum_{\bm{x}_{i} \in \mathcal{D}_{u}}\left\{\frac{1}{\lvert \mathcal{Z}_{i} \rvert} \sum_{\bm{z}_{+} \in \mathcal{Z}_{i}} \mathcal{L}_\text{rep}^\text{sqrt} \left( \bm{p}_{i},\sg(\bm{z}_{+}) \right) \right\}.
  \label{eq:r_risk}
\end{equation}
For each predicted class $c$, $\bm{\mu}_{c}=\frac{1}{\lvert \mathcal{U}_{c} \rvert}\sum_{\bm{x}_{i} \in \mathcal{U}_{c}} t(\bm{x}_{i})$ is the empirical class center, where $\tilde{\bm{p}}_{i}=t(\bm{x}_{i})$.
\begin{theorem}\label{the}
Assume data from the same class in the non-contrastive output space follow a $d$-variate von Mises-Fisher distribution: $\varphi (\bm{x}_{i} \mid \tilde{\bm{\mu}}_c,\kappa)=\mathcal{C}_{d}(\kappa)e^{\kappa \tilde{\bm{\mu}}_{c}^\top t(\bm{x}_{i})}$, where $\tilde{\bm{\mu}}_{c} = \bm{\mu}_{c}/\norm{\bm{\mu}_{c}}$, $\kappa>0$ is a fixed concentration parameter, and $\mathcal{C}_{d}(\kappa)$ is the normalization constant. Then,
\begin{equation}
  \sum_{\mathcal{U}_{c} \in \mathcal{D}_{u}} \kappa {\lvert \mathcal{U}_{c} \rvert}\norm{\bm{\mu}_{c}} \geq \kappa n_{u}-n_{u}\tilde{\mathcal{R}}(\bm{x}).
\end{equation}
\end{theorem}
A detailed proof is provided in~\ref{apx:EM}. Theorem~\ref{the} indicates that minimizing $\tilde{\mathcal{R}}(\bm{x})$ increases a lower bound on the log-likelihood in Eq.~(\ref{eq:log-likelihood}). Further analysis of the gap between $\sum_{\mathcal{U}_{c} \in \mathcal{D}_{u}} \kappa {\lvert \mathcal{U}_{c} \rvert}\norm{\bm{\mu}_{c}}$ and $\kappa n_{u}-n_{u}\tilde{\mathcal{R}}(\bm{x})$ in~\ref{apx:EM} shows that, although strict monotonic tightening of the bound at every iteration is not guaranteed, the lower bound becomes asymptotically tight as the within-class representations become increasingly aligned.

\section{Experiments}\label{sec:experiments}

\subsection{Experimental settings}
\subsubsection{Implementation details}
Optimization was performed using stochastic gradient descent (SGD) with a momentum of 0.9 and a cosine annealing learning-rate schedule. The initial learning rate was set to 0.001 for all disaster events. The model was trained for 400 epochs with a batch size of 32. Unless otherwise specified, $\alpha$ and $k$ were set to 0.99 and 9, respectively. The parameter $\beta$ was set to 0.9 for the hurricane and tornado events and 0.99 for all other disaster events, while $w_{rep}$ was set to 30 for the tsunami event and 10 for all other disaster events. By default, the GFM was initialized with weights pretrained using RemoteCLIP~\citep{10504785} and adapted using deep visual prompt tuning (VPT-deep)~\citep{10.1007/978-3-031-19827-4_41}. During the first 30 epochs, the pseudo-labels were kept fixed as a warm-up strategy before iterative pseudo-target refinement.

\subsubsection{Evaluation metrics}
Overall accuracy (OA) and F1-score were employed as the primary metrics for performance evaluation, while precision (P) and recall (R) were additionally reported in selected experiments for supplementary analysis. OA reflects the overall classification performance, whereas the F1-score provides a more informative assessment under class imbalance, which is common in building damage mapping. Unless otherwise specified, all metrics were computed at the object level. All experiments were repeated three times, with the mean and standard deviation (std) reported.

\subsection{Benchmark methods}
Two categories of methods were considered to evaluate the effectiveness of the proposed PUMA framework: training-from-scratch methods and GFM fine-tuning methods. Training-from-scratch methods do not adapt a pretrained GFM but instead train the model from scratch. Among these methods, nnPU~\citep{kiryo2017positive,ZHAO2022328} and WSC~\citep{zhou2025weaklysupervised} require additional parameters to be estimated prior to training, whereas vPU~\citep{NEURIPS2020_aa0d2a80}, HolisticPU~\citep{xinrui2023beyond}, and NcPU~\citep{zhao2026noisypair} do not rely on such prior estimates. In addition, WSC and NcPU incorporate representation learning into the training process. PUMA was further compared with several methods under the GFM fine-tuning setting. CE-GFM was used as a baseline that fine-tunes the GFM using cross-entropy loss, with all unlabeled samples naively treated as negatives. nnPU was also adapted to the GFM fine-tuning setting, referred to as nnPU-GFM. ROUTE~\citep{li2026weakly} is a foundation model fine-tuning method specifically designed for PU learning; however, it requires additional parameters to be estimated prior to training.

\subsection{Benchmark comparison and analysis}

\begin{table}
\centering
\caption{Quantitative comparison of different methods. The best result for each metric is highlighted in red.}
\label{tab:main_results}
\resizebox{0.98\linewidth}{!}{%
\begin{tblr}{
  row{1} = {c},
  row{2} = {c},
  row{3} = {font=\itshape},
  row{9} = {font=\itshape},
  cell{1}{1} = {r=2}{},
  cell{1}{2} = {c=2}{},
  cell{1}{4} = {c=2}{},
  cell{1}{6} = {c=2}{},
  cell{1}{8} = {c=2}{},
  cell{3}{1} = {c=9}{},
  cell{4}{2} = {c},
  cell{4}{3} = {c},
  cell{4}{4} = {c},
  cell{4}{5} = {c},
  cell{4}{6} = {c},
  cell{4}{7} = {c},
  cell{4}{8} = {c},
  cell{4}{9} = {c},
  cell{5}{2} = {c},
  cell{5}{3} = {c},
  cell{5}{4} = {c},
  cell{5}{5} = {c},
  cell{5}{6} = {c},
  cell{5}{7} = {c},
  cell{5}{8} = {c},
  cell{5}{9} = {c},
  cell{6}{2} = {c},
  cell{6}{3} = {c},
  cell{6}{4} = {c},
  cell{6}{5} = {c},
  cell{6}{6} = {c},
  cell{6}{7} = {c},
  cell{6}{8} = {c},
  cell{6}{9} = {c},
  cell{7}{2} = {c},
  cell{7}{3} = {c},
  cell{7}{4} = {c},
  cell{7}{5} = {c},
  cell{7}{6} = {c},
  cell{7}{7} = {c},
  cell{7}{8} = {c},
  cell{7}{9} = {c},
  cell{8}{2} = {c},
  cell{8}{3} = {c},
  cell{8}{4} = {c},
  cell{8}{5} = {c},
  cell{8}{6} = {c,fg=red},
  cell{8}{7} = {c,fg=red},
  cell{8}{8} = {c},
  cell{8}{9} = {c},
  cell{9}{1} = {c=9}{},
  cell{10}{2} = {c},
  cell{10}{3} = {c},
  cell{10}{4} = {c},
  cell{10}{5} = {c},
  cell{10}{6} = {c},
  cell{10}{7} = {c},
  cell{10}{8} = {c},
  cell{10}{9} = {c},
  cell{11}{2} = {c},
  cell{11}{3} = {c},
  cell{11}{4} = {c},
  cell{11}{5} = {c},
  cell{11}{6} = {c},
  cell{11}{7} = {c},
  cell{11}{8} = {c},
  cell{11}{9} = {c},
  cell{12}{2} = {c},
  cell{12}{3} = {c},
  cell{12}{4} = {c},
  cell{12}{5} = {c},
  cell{12}{6} = {c},
  cell{12}{7} = {c},
  cell{12}{8} = {c},
  cell{12}{9} = {c},
  cell{13}{2} = {c},
  cell{13}{3} = {c},
  cell{13}{4} = {c,fg=red},
  cell{13}{5} = {c,fg=red},
  cell{13}{6} = {c},
  cell{13}{7} = {c},
  cell{13}{8} = {c,fg=red},
  cell{13}{9} = {c,fg=red},
  cell{14}{2} = {c,fg=red},
  cell{14}{3} = {c,fg=red},
  cell{14}{4} = {c},
  cell{14}{5} = {c},
  cell{14}{6} = {c},
  cell{14}{7} = {c},
  cell{14}{8} = {c},
  cell{14}{9} = {c},
  hline{1,3-4,9-10,15} = {-}{},
  hline{2} = {2-9}{},
}
Method                                                   & Tsunami                      &                               & Hurricane                     &                               & Tornado                       &                               & Wildfire                      &                               \\
                                                         & OA                           & F1                            & OA                            & F1                            & OA                            & F1                            & OA                            & F1                            \\
\textit{\textit{\textit{Training-from-scratch methods}}} &                              &                               &                               &                               &                               &                               &                               &                               \\
nnPU                                                     & 78.48\textsuperscript{±0.55} & 81.84\textsuperscript{±0.38}  & 92.76\textsuperscript{±0.52}  & 82.57\textsuperscript{±1.13}  & 95.60\textsuperscript{±0.50}  & 92.87\textsuperscript{±0.74}  & 93.92\textsuperscript{±0.06}  & 87.27\textsuperscript{±0.18}  \\
vPU                                                      & 65.82\textsuperscript{±5.39} & 48.77\textsuperscript{±12.71} & 84.62\textsuperscript{±1.91}  & 27.85\textsuperscript{±14.66} & 72.42\textsuperscript{±1.75}  & 13.79\textsuperscript{±10.01} & 80.65\textsuperscript{±2.43}  & 27.22\textsuperscript{±15.78} \\
HolisticPU                                               & 80.67\textsuperscript{±1.89} & 78.50\textsuperscript{±2.85}  & 51.90\textsuperscript{±30.25} & 43.68\textsuperscript{±16.45} & 82.16\textsuperscript{±18.10} & 77.98\textsuperscript{±16.52} & 56.19\textsuperscript{±33.00} & 54.72\textsuperscript{±28.19} \\
WSC                                                      & 86.41\textsuperscript{±0.95} & 85.83\textsuperscript{±1.21}  & 92.59\textsuperscript{±1.23}  & 81.97\textsuperscript{±2.20}  & 95.69\textsuperscript{±0.72}  & 92.92\textsuperscript{±1.02}  & 95.05\textsuperscript{±0.26}  & 89.78\textsuperscript{±0.37}  \\
NcPU                                                     & 86.80\textsuperscript{±0.42} & 86.49\textsuperscript{±0.34}  & 94.17\textsuperscript{±1.70}  & 85.11\textsuperscript{±3.38}  & 95.75\textsuperscript{±0.18}  & 92.94\textsuperscript{±0.30}  & 94.55\textsuperscript{±0.99}  & 88.33\textsuperscript{±1.88}  \\
\textit{\textit{\textit{GFM fine-tuning methods}}}       &                              &                               &                               &                               &                               &                               &                               &                               \\
CE-GFM                                                   & 62.21\textsuperscript{±1.31} & 39.91\textsuperscript{±3.19}  & 90.79\textsuperscript{±0.76}  & 68.45\textsuperscript{±3.29}  & 83.65\textsuperscript{±1.87}  & 64.12\textsuperscript{±5.50}  & 88.56\textsuperscript{±1.59}  & 67.38\textsuperscript{±5.98}  \\
nnPU-GFM                                                 & 84.54\textsuperscript{±2.13} & 86.40\textsuperscript{±1.61}  & 95.84\textsuperscript{±0.38}  & 89.25\textsuperscript{±0.99}  & 95.46\textsuperscript{±0.25}  & 92.45\textsuperscript{±0.36}  & 97.02\textsuperscript{±0.17}  & 93.45\textsuperscript{±0.41}  \\
ROUTE                                                    & 86.30\textsuperscript{±0.24} & 87.73\textsuperscript{±0.21}  & 86.84\textsuperscript{±4.40}  & 73.80\textsuperscript{±6.47}  & 92.47\textsuperscript{±0.67}  & 88.62\textsuperscript{±0.93}  & 96.59\textsuperscript{±0.29}  & 92.79\textsuperscript{±0.50}  \\
PUMA ($\mathcal{L}_\text{rep}^\text{log}$)               & 93.40\textsuperscript{±0.30} & 93.57\textsuperscript{±0.29}  & 97.65\textsuperscript{±0.28}  & 93.72\textsuperscript{±0.73}  & 95.72\textsuperscript{±0.29}  & 92.88\textsuperscript{±0.46}  & 98.07\textsuperscript{±0.17}  & 95.79\textsuperscript{±0.34}  \\
PUMA ($\mathcal{L}_\text{rep}^\text{sqrt}$)              & 93.63\textsuperscript{±0.32} & 93.71\textsuperscript{±0.33}  & 97.32\textsuperscript{±0.60}  & 92.73\textsuperscript{±1.80}  & 95.66\textsuperscript{±0.27}  & 92.77\textsuperscript{±0.38}  & 98.01\textsuperscript{±0.04}  & 95.65\textsuperscript{±0.08}  
\end{tblr}
}
\end{table}

\begin{figure}[!t]
\centering
\includegraphics[width=0.98\linewidth]{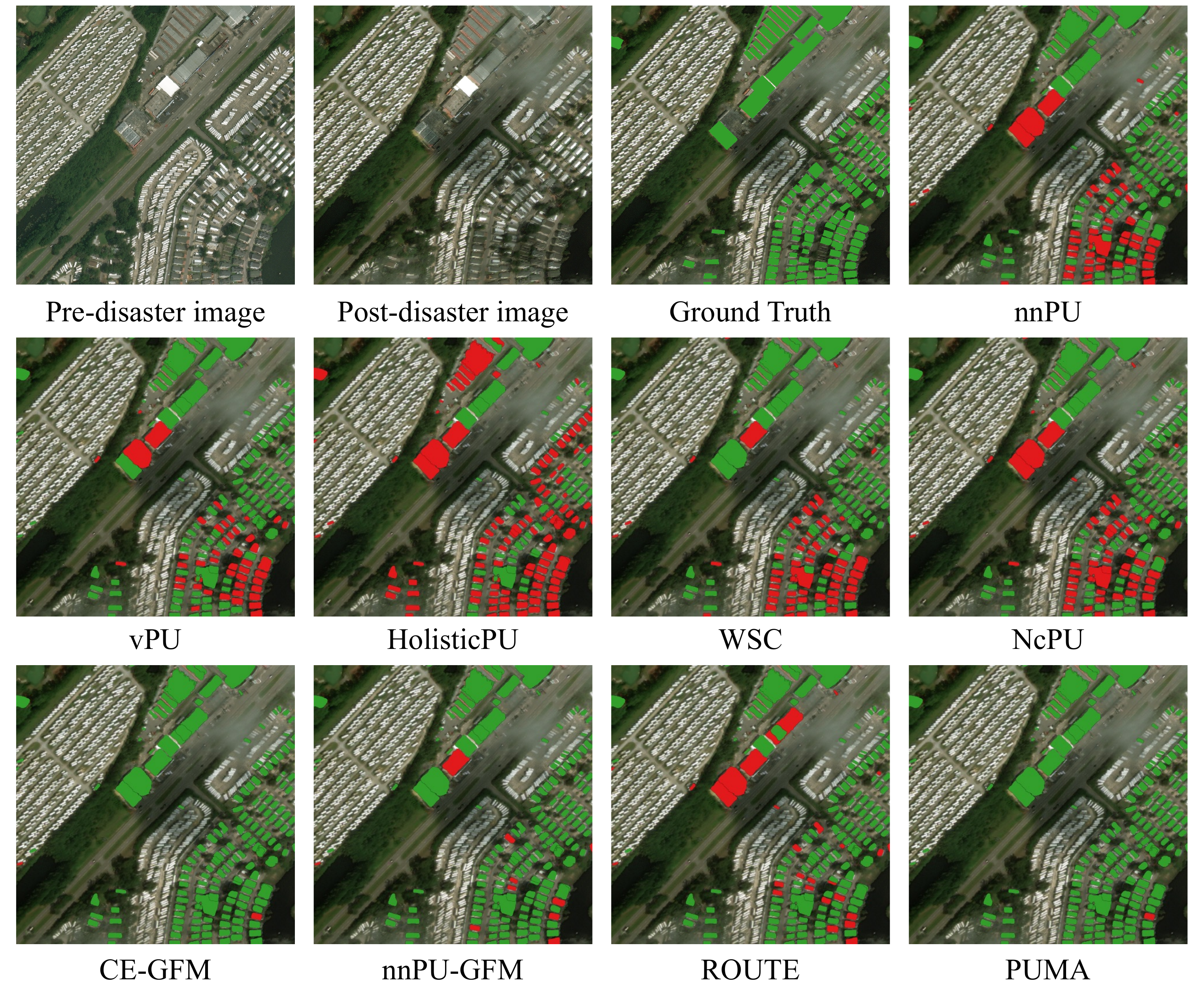}
\caption{Visual comparison of different methods on the Hurricane Florence subset of the xBD dataset. Damaged and undamaged buildings are shown in \textcolor{red}{red} and \textcolor{darkgreen}{green}, respectively.}
\label{fig:mapping_results_hurricane}
\end{figure}

\begin{figure}[!t]
\centering
\includegraphics[width=0.98\linewidth]{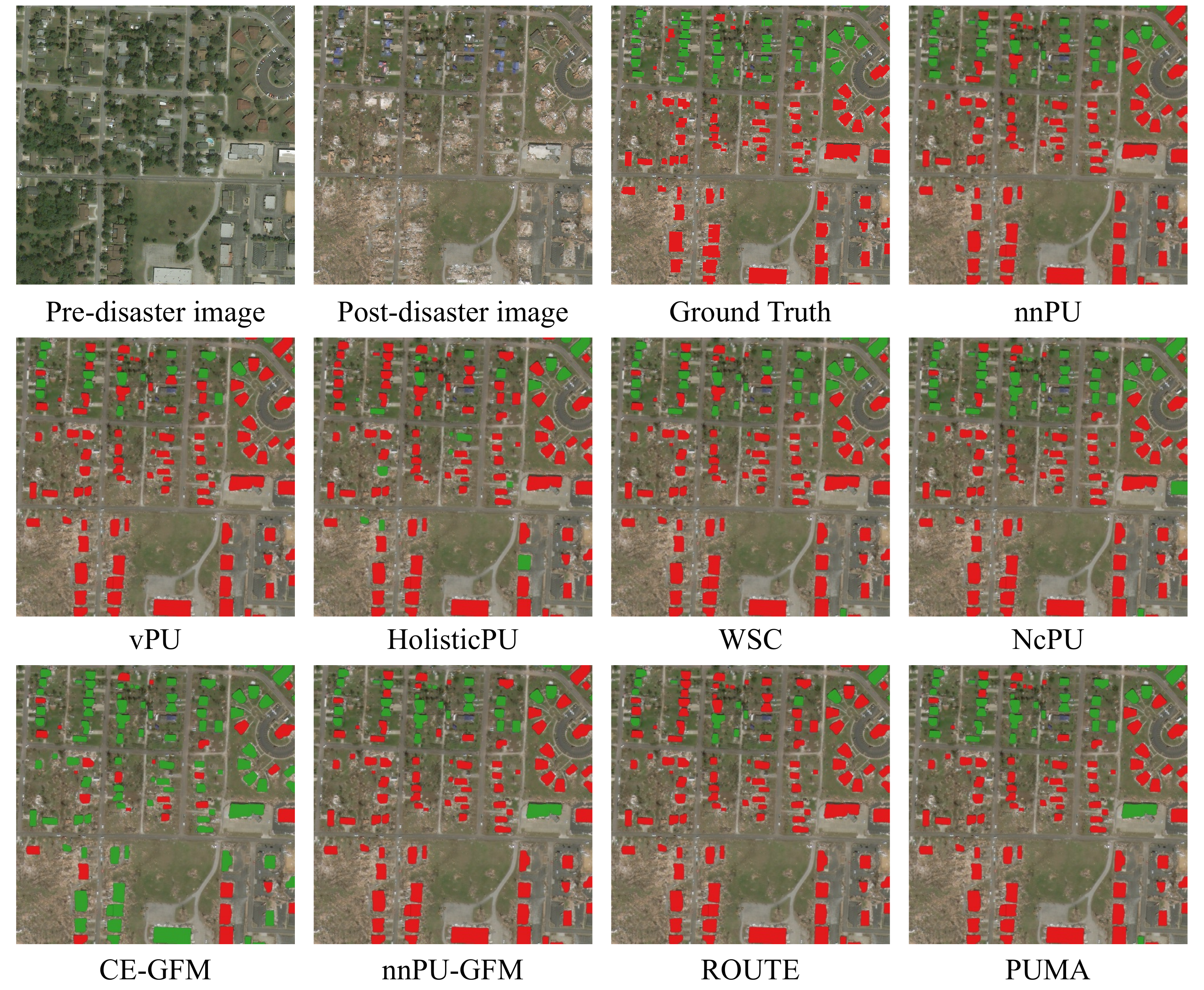}
\caption{Visual comparison of different methods on the Joplin Tornado subset of the xBD dataset. Damaged and undamaged buildings are shown in \textcolor{red}{red} and \textcolor{darkgreen}{green}, respectively.}
\label{fig:mapping_results_tornado}
\end{figure}

\noindent \textbf{PUMA achieves the best overall performance}. The quantitative results of the different methods are summarized in Table~\ref{tab:main_results}, while qualitative comparisons are presented in Figure~\ref{fig:mapping_results_hurricane} and Figure~\ref{fig:mapping_results_tornado}. Compared with both training-from-scratch methods and GFM fine-tuning methods, PUMA demonstrated consistently strong performance across different disaster types. Specifically, PUMA achieved the highest OA and F1-score for the tsunami, hurricane, and wildfire events. Relative to the strongest competing methods, PUMA yielded clear performance gains, particularly for the tsunami and hurricane events, with F1-score improvements of 5.98 and 4.47 percentage points, respectively. For the tornado event, PUMA achieved an OA of 95.72\% and an F1-score of 92.88\%, remaining highly competitive with the best-performing NcPU. Notably, PUMA consistently and substantially outperformed conventional CE-GFM across all disaster events, indicating that directly treating unlabeled samples as negative samples is suboptimal under the PU learning setting. Moreover, PUMA consistently outperformed nnPU-GFM and ROUTE in GFM adaptation, demonstrating its effectiveness in leveraging limited positive annotations together with unlabeled samples for weakly supervised GFM fine-tuning.

\noindent \textbf{PUMA learns more discriminative representations.} The t-SNE visualizations of the representations extracted from the test data (Figure~\ref{fig:representation_overlap}) further illustrate the representation learning capabilities of different methods under unreliable supervision. Compared with the other GFM fine-tuning methods, PUMA yielded more compact intra-class distributions and clearer inter-class separation between damaged and undamaged buildings, indicating that it learns more discriminative representations. These results further underscore the importance of effective representation learning in the foundation-model fine-tuning paradigm, particularly when supervision is unreliable.

\noindent \textbf{PUMA learns more accurate decision boundaries.} Unlike existing GFM fine-tuning methods that rely on a pre-estimated class prior, PUMA requires no prior information during fine-tuning, thereby avoiding the propagation of errors arising from inaccurate class-prior estimation. Consequently, PUMA learned more accurate decision boundaries between damaged and undamaged buildings, as illustrated in Figure~\ref{fig:representation_overlap}.

\subsection{Ablation and methodological analysis}
\subsubsection{Ablation analysis}

\begin{table}
\centering
\caption{Ablation study of the proposed modules. The parameter $k$ in $\mathcal{L}_\text{rep}^\text{sqrt}$ was set to 9 by default.}
\label{tab:abla_results}
\resizebox{0.98\linewidth}{!}{%
\begin{tblr}{
  cells = {c},
  cell{1}{1} = {r=2}{},
  cell{1}{2} = {r=2}{},
  cell{1}{3} = {c=4}{},
  cell{1}{7} = {c=4}{},
  hline{1,3,8-9,13} = {-}{},
  hline{2} = {3-10}{},
}
{Representation\\Learning}                   & {Pseudo-label\\Updating} & Tsunami                      &                               &                              &                               & Hurricane                     &                               &                               &                              \\
                                           &                          & OA                           & F1                            & P                            & R                             & OA                            & F1                            & P                             & R                            \\
$\mathcal{L}_\text{rep}^\text{self}$       &                          & 68.70\textsuperscript{±1.90} & 54.83\textsuperscript{±3.99}  & 97.66\textsuperscript{±0.83} & 38.18\textsuperscript{±3.76}  & 91.39\textsuperscript{±0.65}  & 70.24\textsuperscript{±3.00}  & 96.87\textsuperscript{±0.24}  & 55.16\textsuperscript{±3.80} \\
$\mathcal{L}_\text{rep}^\text{sup}$        &                          & 65.32\textsuperscript{±8.56} & 45.60\textsuperscript{±18.94} & 97.78\textsuperscript{±0.25} & 31.22\textsuperscript{±17.58} & 92.72\textsuperscript{±1.03}  & 76.10\textsuperscript{±4.53}  & 96.18\textsuperscript{±1.05}  & 63.17\textsuperscript{±6.44} \\
$\mathcal{L}_\text{rep}^\text{sqrt} (k=1)$ &                          & 67.14\textsuperscript{±4.50} & 51.02\textsuperscript{±10.06} & 97.73\textsuperscript{±0.67} & 34.92\textsuperscript{±8.99}  & 93.53\textsuperscript{±0.13}  & 79.36\textsuperscript{±0.35}  & 96.58\textsuperscript{±0.75}  & 67.35\textsuperscript{±0.15} \\
$\mathcal{L}_\text{rep}^\text{log}$        &                          & 91.15\textsuperscript{±0.85} & 90.57\textsuperscript{±1.01}  & 96.54\textsuperscript{±0.75} & 85.32\textsuperscript{±1.90}  & 94.17\textsuperscript{±0.71}  & 81.87\textsuperscript{±2.82}  & 95.95\textsuperscript{±1.24}  & 71.53\textsuperscript{±4.86} \\
$\mathcal{L}_\text{rep}^\text{sqrt}$       &                          & 85.59\textsuperscript{±6.10} & 83.00\textsuperscript{±8.01}  & 97.99\textsuperscript{±0.82} & 72.68\textsuperscript{±13.12} & 93.86\textsuperscript{±1.01}  & 80.54\textsuperscript{±4.05}  & 96.52\textsuperscript{±1.54}  & 69.31\textsuperscript{±6.33} \\
                                           & $s$                      & 91.48\textsuperscript{±0.46} & 91.36\textsuperscript{±0.36}  & 92.58\textsuperscript{±3.22} & 90.31\textsuperscript{±2.96}  & 96.61\textsuperscript{±0.16}  & 90.37\textsuperscript{±0.48}  & 95.19\textsuperscript{±0.25}  & 86.02\textsuperscript{±0.90} \\
$\mathcal{L}_\text{rep}^\text{sqrt}$       & $\tilde{s}$              & 92.75\textsuperscript{±0.39} & 93.00\textsuperscript{±0.34}  & 89.74\textsuperscript{±0.80} & 96.50\textsuperscript{±0.43}  & 86.58\textsuperscript{±17.16} & 78.05\textsuperscript{±22.27} & 70.52\textsuperscript{±30.56} & 95.82\textsuperscript{±3.34} \\
$\mathcal{L}_\text{rep}^\text{sqrt}$       & $\tilde{s}$+SAT          & 89.87\textsuperscript{±1.68} & 88.97\textsuperscript{±2.01}  & 97.26\textsuperscript{±0.83} & 82.02\textsuperscript{±3.33}  & 96.38\textsuperscript{±0.29}  & 89.64\textsuperscript{±0.83}  & 95.04\textsuperscript{±0.82}  & 84.82\textsuperscript{±1.03} \\
$\mathcal{L}_\text{rep}^\text{log}$        & $s$                      & 93.40\textsuperscript{±0.30} & 93.57\textsuperscript{±0.29}  & 91.06\textsuperscript{±0.50} & 96.22\textsuperscript{±0.48}  & 97.65\textsuperscript{±0.28}  & 93.72\textsuperscript{±0.73}  & 92.68\textsuperscript{±1.12}  & 94.80\textsuperscript{±1.06} \\
$\mathcal{L}_\text{rep}^\text{sqrt}$       & $s$                      & 93.63\textsuperscript{±0.32} & 93.71\textsuperscript{±0.33}  & 92.41\textsuperscript{±0.53} & 95.05\textsuperscript{±0.84}  & 97.32\textsuperscript{±0.60}  & 92.73\textsuperscript{±1.80}  & 92.77\textsuperscript{±0.80}  & 92.75\textsuperscript{±3.91} 
\end{tblr}
}
\end{table}

\noindent \textbf{$\mathcal{L}_\text{rep}^\text{sqrt}$ and $\mathcal{L}_\text{rep}^\text{log}$ demonstrate greater robustness to unreliable supervision.} As shown in Table~\ref{tab:abla_results}, $\mathcal{L}_\text{rep}^\text{sqrt}$ and $\mathcal{L}_\text{rep}^\text{log}$ outperformed $\mathcal{L}_\text{rep}^\text{sup}$ in terms of both OA and F1-score, demonstrating the robustness of the representation refinement losses to noisy pairs. Further analysis shows that these improvements are primarily driven by the higher recall of damaged buildings, indicating that $\mathcal{L}_\text{rep}^\text{sqrt}$ and $\mathcal{L}_\text{rep}^\text{log}$ are more robust to unreliable supervision.

\noindent \textbf{NeWER and LaTTE mutually reinforce each other.} As shown in Table~\ref{tab:abla_results}, neither $\mathcal{L}_\text{rep}^\text{sqrt}$ alone nor $s$ alone achieved the best performance, whereas their joint use consistently yielded improved results. Specifically, $\mathcal{L}_\text{rep}^\text{sqrt}$ facilitates label disambiguation in LaTTE by enhancing the discriminability of the learned representations, while the refined supervision provided by $s$ improves the reliability of the supervision used for non-contrastive representation learning in NeWER. This mutually reinforcing mechanism can be interpreted within a GEM-inspired framework, as detailed in Section~\ref{sec:EM}.

\noindent \textbf{The latent-state-based conservative pseudo-label refinement plays a critical role in LaTTE.} In the absence of explicit supervision for the undamaged class, $\tilde{s}$ may yield deceptively high recall at the expense of low precision, as observed for the hurricane event. Introducing explicit negative supervision through SAT can improve precision; however, it may also result in a substantial reduction in recall, as observed for the tsunami event. This degradation can be attributed to the accumulation of erroneous negative supervision during fine-tuning, where incorrectly assigned negative labels progressively reinforce subsequent prediction errors. In contrast, $s$ introduces explicit supervision for the undamaged class while preventing erroneous negative supervision from accumulating across iterations, thereby achieving a more favorable balance between precision and recall and ultimately improving overall performance.

\subsubsection{$\mathcal{L}_\text{rep}^\text{sqrt}$ enhances existing weakly supervised methods for GFM fine-tuning}

\begin{table}
\centering
\caption{Effect of representation refinement loss on existing PU learning methods.}
\label{tab:risk_Lrep}
\resizebox{0.55\linewidth}{!}{%
\begin{tblr}{
  row{1} = {c},
  cell{2}{2} = {c},
  cell{2}{3} = {c},
  cell{2}{4} = {c},
  cell{2}{5} = {c},
  cell{3}{2} = {c},
  cell{3}{3} = {c},
  cell{3}{4} = {c},
  cell{3}{5} = {c},
  cell{4}{2} = {c},
  cell{4}{3} = {c},
  cell{4}{4} = {c},
  cell{4}{5} = {c},
  cell{5}{2} = {c},
  cell{5}{3} = {c},
  cell{5}{4} = {c},
  cell{5}{5} = {c},
  hline{1-2,6} = {-}{},
}
Method                         & OA                           & F1                           & P                            & R                            \\
uPU                            & 83.48\textsuperscript{±2.31} & 81.73\textsuperscript{±3.15} & 90.88\textsuperscript{±1.86} & 74.40\textsuperscript{±5.32} \\
uPU+$\mathcal{L}_\text{rep}^\text{sqrt}$  & 89.34\textsuperscript{±0.31} & 89.11\textsuperscript{±0.58} & 90.85\textsuperscript{±1.78} & 87.52\textsuperscript{±2.78} \\
nnPU                           & 88.18\textsuperscript{±1.82} & 87.48\textsuperscript{±2.07} & 92.67\textsuperscript{±1.04} & 82.87\textsuperscript{±3.12} \\
nnPU+$\mathcal{L}_\text{rep}^\text{sqrt}$ & 90.03\textsuperscript{±0.24} & 90.10\textsuperscript{±0.46} & 89.25\textsuperscript{±1.54} & 91.04\textsuperscript{±2.50} 
\end{tblr}
}
\end{table}

Risk-estimation-based methods have been theoretically and empirically validated as simple yet robust solutions to weakly supervised learning problems. As shown in Table~\ref{tab:risk_Lrep}, incorporating $\mathcal{L}_\text{rep}^\text{sqrt}$ consistently improved the performance of both uPU~\citep{pmlr-v37-plessis15} and nnPU~\citep{kiryo2017positive,ZHAO2022328} for GFM adaptation in the tsunami event. These results further highlight the importance of learning discriminative representations for effective GFM fine-tuning.

\subsubsection{Compatibility of PUMA with different PEFT methods}

\begin{table}[!t]
\centering
\caption{Performance of PUMA with different PEFT strategies.}
\label{tab:peft_results}
\resizebox{0.65\linewidth}{!}{%
\begin{tabular}{llcccc} 
\hline
\multicolumn{2}{c}{\multirow{2}{*}{Method}} & \multicolumn{2}{c}{Tsunami}                                 & \multicolumn{2}{c}{Hurricane}                                  \\ 
\cline{3-6}
\multicolumn{2}{c}{}                         & OA                           & F1                           & OA                            & F1                             \\ 
\hline
\multicolumn{2}{l}{Full fine-tuning}         & 50.45\textsuperscript{±0.95} & 66.44\textsuperscript{±0.23} & 46.64\textsuperscript{±39.64} & 48.14\textsuperscript{±28.41}  \\
\multicolumn{2}{l}{Linear probing}   & 90.80\textsuperscript{±0.26} & 91.12\textsuperscript{±0.26} & 94.57\textsuperscript{±0.75}  & 86.53\textsuperscript{±1.56}   \\ 
\hline
\multirow{6}{*}{PEFT     } & BitFit          & 92.92\textsuperscript{±0.84} & 93.07\textsuperscript{±0.73} & 97.34\textsuperscript{±0.07}  & 92.86\textsuperscript{±0.29}   \\
                           & VPT-shallow     & 90.83\textsuperscript{±0.70} & 91.11\textsuperscript{±0.57} & 95.81\textsuperscript{±0.15}  & 89.35\textsuperscript{±0.41}   \\
                           & VPT-deep        & 93.63\textsuperscript{±0.32} & 93.71\textsuperscript{±0.33} & 97.32\textsuperscript{±0.60}  & 92.73\textsuperscript{±1.80}   \\
                           & Adapter         & 92.73\textsuperscript{±1.35} & 92.91\textsuperscript{±1.17} & 97.07\textsuperscript{±0.17}  & 92.00\textsuperscript{±0.44}   \\
                           & AdaptFormer     & 93.65\textsuperscript{±0.81} & 93.68\textsuperscript{±0.78} & 97.12\textsuperscript{±0.76}  & 92.16\textsuperscript{±2.18}   \\
                           & LoRA            & 92.38\textsuperscript{±1.06} & 92.28\textsuperscript{±1.05} & 97.21\textsuperscript{±0.38}  & 92.35\textsuperscript{±1.15}   \\
\hline
\end{tabular}
}
\end{table}

PUMA exhibits strong compatibility with different PEFT methods. As shown in Table~\ref{tab:peft_results}, integrating PUMA with various PEFT strategies, including BitFit~\citep{ben-zaken-etal-2022-bitfit}, VPT~\citep{10.1007/978-3-031-19827-4_41}, Adapter~\citep{houlsby2019parameter}, AdaptFormer~\citep{chen2022adaptformer}, and LoRA~\citep{hu2022lora}, consistently yielded strong performance across both disaster events. Most PEFT variants achieved OA and F1-scores above 90\% for both the tsunami and hurricane events, demonstrating that the effectiveness of PUMA is not dependent on a specific PEFT strategy. In contrast, full fine-tuning led to substantially degraded performance and greater performance variability, particularly for the hurricane event, suggesting that updating all GFM parameters can be unstable under weak supervision. Linear probing, i.e., classifier-only fine-tuning, provided a considerably stronger baseline but remained inferior to most PEFT-based variants, especially in terms of F1-score for the hurricane event. Among the evaluated strategies, VPT-deep achieved the strongest overall performance. These results demonstrate that PUMA can be flexibly integrated with a broad range of PEFT methods while maintaining consistently competitive performance, highlighting its robustness and general applicability across different GFM adaptation strategies.

\subsubsection{Results of PUMA with different pre-trained models}

\begin{table}
\centering
\caption{Performance of PUMA with different GFMs.}
\label{tab:prtr_results}
\resizebox{0.65\linewidth}{!}{%
\begin{tblr}{
  row{1} = {c},
  row{2} = {c},
  row{5} = {c},
  row{8} = {c},
  cell{1}{1} = {r=2}{},
  cell{1}{2} = {r=2}{},
  cell{1}{3} = {c=2}{},
  cell{1}{5} = {c=2}{},
  cell{3}{2} = {c},
  cell{3}{3} = {c},
  cell{3}{4} = {c},
  cell{3}{5} = {c},
  cell{3}{6} = {c},
  cell{4}{1} = {r=2}{},
  cell{4}{2} = {c},
  cell{4}{3} = {c},
  cell{4}{4} = {c},
  cell{4}{5} = {c},
  cell{4}{6} = {c},
  cell{6}{2} = {c},
  cell{6}{3} = {c},
  cell{6}{4} = {c},
  cell{6}{5} = {c},
  cell{6}{6} = {c},
  cell{7}{1} = {r=2}{},
  cell{7}{2} = {c},
  cell{7}{3} = {c},
  cell{7}{4} = {c},
  cell{7}{5} = {c},
  cell{7}{6} = {c},
  cell{9}{2} = {c},
  cell{9}{3} = {c},
  cell{9}{4} = {c},
  cell{9}{5} = {c},
  cell{9}{6} = {c},
  cell{10}{2} = {c},
  cell{10}{3} = {c},
  cell{10}{4} = {c},
  cell{10}{5} = {c},
  cell{10}{6} = {c},
  hline{1,3,11} = {-}{},
  hline{2} = {3-6}{},
}
{Pre-trained\\model} & Backbone  & Tsunami                      &                              & Hurricane                    &                              \\
                     &           & OA                           & F1                           & OA                           & F1                           \\
scratch              & ViT-B     & 80.98\textsuperscript{±2.25} & 83.34\textsuperscript{±1.44} & 91.20\textsuperscript{±1.90} & 80.01\textsuperscript{±3.22} \\
CLIP                 & ResNet-50 & 91.18\textsuperscript{±0.09} & 91.39\textsuperscript{±0.37} & 96.95\textsuperscript{±0.69} & 92.07\textsuperscript{±1.57} \\
                     & ViT-B     & 93.79\textsuperscript{±0.57} & 93.85\textsuperscript{±0.49} & 97.53\textsuperscript{±0.30} & 93.26\textsuperscript{±0.81} \\
LCLIP                & ViT-B     & 93.76\textsuperscript{±0.75} & 93.83\textsuperscript{±0.69} & 97.46\textsuperscript{±0.29} & 93.12\textsuperscript{±0.86} \\
RemoteCLIP           & ResNet-50 & 93.44\textsuperscript{±0.81} & 93.49\textsuperscript{±0.70} & 97.40\textsuperscript{±0.26} & 93.11\textsuperscript{±0.69} \\
                     & ViT-B     & 93.63\textsuperscript{±0.32} & 93.71\textsuperscript{±0.33} & 97.32\textsuperscript{±0.60} & 92.73\textsuperscript{±1.80} \\
GeoRSCLIP            & ViT-B     & 94.06\textsuperscript{±0.24} & 94.08\textsuperscript{±0.23} & 97.45\textsuperscript{±0.17} & 93.14\textsuperscript{±0.47} \\
LRSCLIP              & ViT-B     & 94.02\textsuperscript{±0.45} & 93.96\textsuperscript{±0.40} & 96.95\textsuperscript{±0.09} & 91.57\textsuperscript{±0.40} 
\end{tblr}
}
\end{table}

\begin{table}
\centering
\caption{Building localization and damage classification performance using automatically extracted building footprints.}
\label{tab:localization_damage}
\resizebox{0.98\linewidth}{!}{%
\begin{tblr}{
  row{1} = {c},
  row{2} = {c},
  cell{1}{2} = {c=3}{},
  cell{1}{5} = {c=6}{},
  cell{3}{2} = {c},
  cell{3}{3} = {c},
  cell{3}{4} = {c},
  cell{3}{5} = {c},
  cell{3}{6} = {c},
  cell{3}{7} = {c},
  cell{3}{8} = {c},
  cell{3}{9} = {c},
  cell{3}{10} = {c},
  cell{4}{2} = {c},
  cell{4}{3} = {c},
  cell{4}{4} = {c},
  cell{4}{5} = {c},
  cell{4}{6} = {c},
  cell{4}{7} = {c},
  cell{4}{8} = {c},
  cell{4}{9} = {c},
  cell{4}{10} = {c},
  cell{5}{2} = {c},
  cell{5}{3} = {c},
  cell{5}{4} = {c},
  cell{5}{5} = {c},
  cell{5}{6} = {c},
  cell{5}{7} = {c},
  cell{5}{8} = {c},
  cell{5}{9} = {c},
  cell{5}{10} = {c},
  hline{1,3,6} = {-}{},
  hline{2} = {2-10}{},
}
Disaster  & Building Localization      &                            &                            & Building Damage Classification &                              &                              &                              &                              &                              \\
Event     & AP\textsuperscript{Box}    & AP\textsuperscript{Mask}   & IoU                        & P\textsuperscript{Pixel}       & R\textsuperscript{Pixel}     & F1\textsuperscript{Pixel}    & P                            & R                            & F1                           \\
Hurricane & 56.7\textsuperscript{±0.2} & 52.2\textsuperscript{±0.1} & 77.8\textsuperscript{±0.6} & 79.75\textsuperscript{±0.26}   & 88.97\textsuperscript{±0.26} & 84.11\textsuperscript{±0.26} & 75.89\textsuperscript{±0.34} & 84.63\textsuperscript{±0.38} & 80.02\textsuperscript{±0.36} \\
Tornado   & 52.1\textsuperscript{±0.4} & 48.2\textsuperscript{±0.2} & 83.0\textsuperscript{±0.4} & 84.26\textsuperscript{±0.14}   & 91.99\textsuperscript{±0.10} & 87.96\textsuperscript{±0.12} & 83.38\textsuperscript{±0.25} & 76.88\textsuperscript{±0.23} & 80.00\textsuperscript{±0.24} \\
Wildfire  & 52.1\textsuperscript{±0.3} & 47.3\textsuperscript{±0.2} & 72.9\textsuperscript{±2.5} & 80.58\textsuperscript{±0.16}   & 89.71\textsuperscript{±0.12} & 84.90\textsuperscript{±0.15} & 80.61\textsuperscript{±0.08} & 84.09\textsuperscript{±0.08} & 82.32\textsuperscript{±0.08} 
\end{tblr}
}
\end{table}

The performance of PUMA with different GFMs, including CLIP~\citep{radford2021learning}, LCLIP~\citep{muhtar2024lhrs}, RemoteCLIP~\citep{10504785}, GeoRSCLIP~\citep{10679571}, and LRSCLIP~\citep{chen2025dgtrsddgtrsclipdualgranularity}, was further evaluated. As shown in Table~\ref{tab:prtr_results}, PUMA achieved consistently competitive performance across different pre-trained models. Compared with training from scratch, initialization from pre-trained models yielded substantial performance improvements. Notably, models pre-trained on natural images (CLIP and LCLIP) achieved overall performance comparable to that of remote sensing-specific pre-trained models (RemoteCLIP, GeoRSCLIP, and LRSCLIP). These results suggest that PEFT methods capable of learning discriminative representations can partially compensate for the limited domain-specific knowledge of general-purpose foundation models, thereby facilitating the effective transfer of models pre-trained on natural images to downstream tasks across different domains.

\subsubsection{Results using automatically extracted building footprints}

\begin{figure}[!t]
\centering
\includegraphics[width=0.98\linewidth]{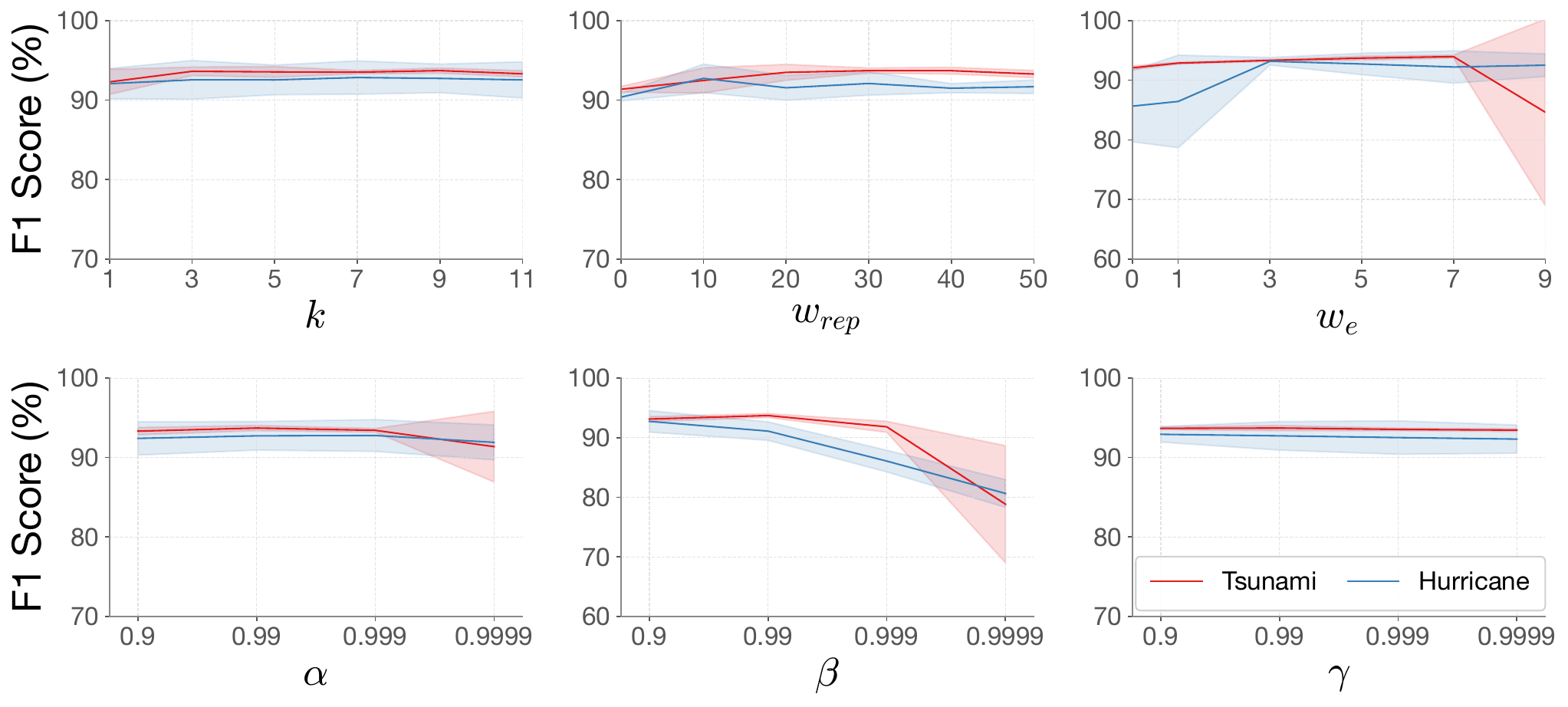}
\caption{Hyperparameter analyses of PUMA for the tsunami and hurricane events.}
\label{fig:hyperparameters}
\end{figure}

\begin{figure}[!t]
\centering
\includegraphics[width=0.98\linewidth]{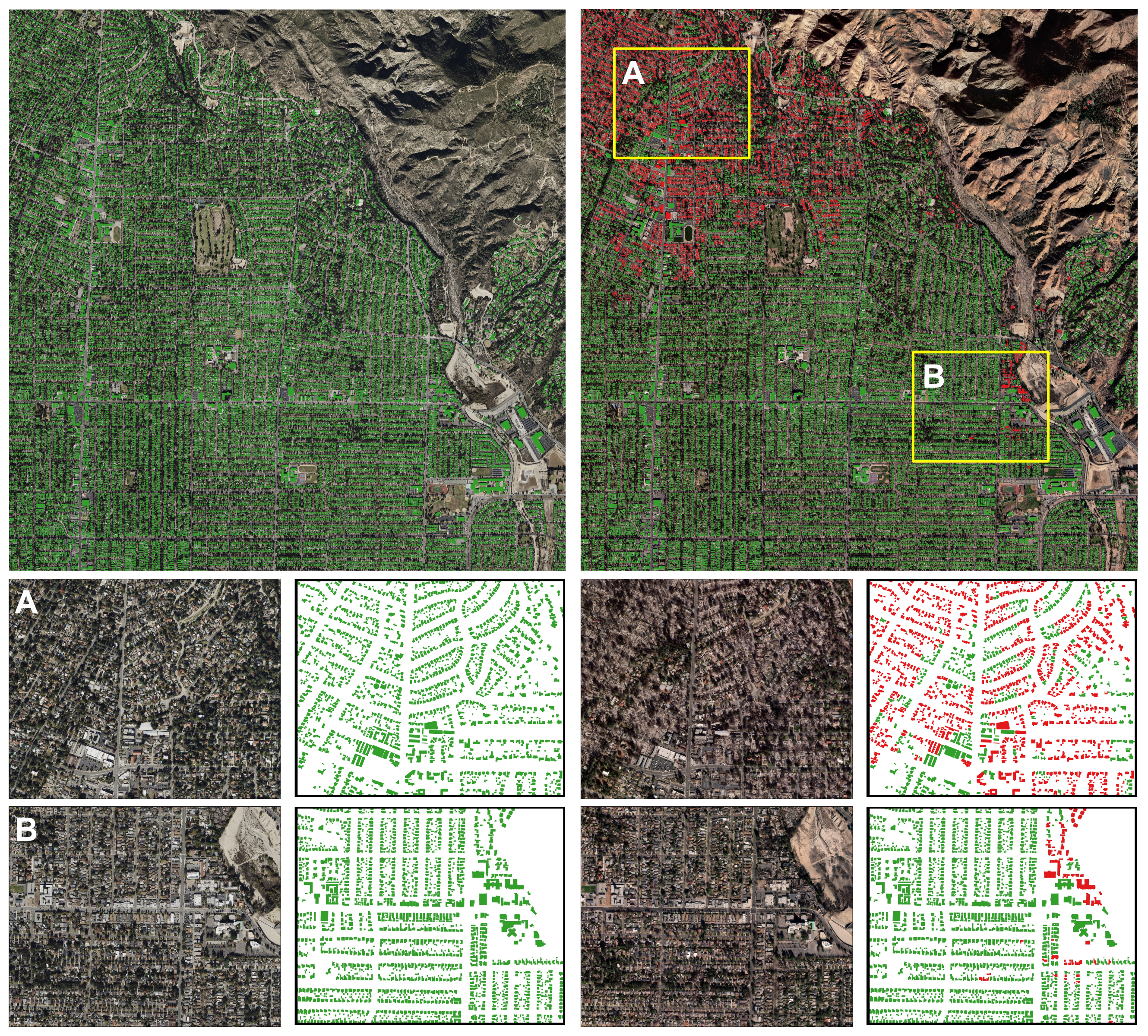}
\caption{Building damage mapping in the Eaton subarea of the Los Angeles wildfires. The left column shows pre-disaster imagery with building footprints, and the right column shows post-disaster imagery with the corresponding damage mapping results. Damaged and undamaged buildings are shown in \textcolor{red}{red} and \textcolor{darkgreen}{green}, respectively.}
\label{fig:local_wildfire_1}
\end{figure}

To further evaluate the applicability of the proposed framework in scenarios where historical building footprints are unavailable, a comprehensive experiment was conducted using automatically extracted building footprints. Specifically, building footprints were generated using the RPN+SAM described in Section~\ref{sec:building_localization}. Building localization performance was evaluated using bounding-box-level average precision (AP\textsuperscript{Box}), mask-level average precision (AP\textsuperscript{Mask}), and intersection over union (IoU), while subsequent building damage classification was assessed using P, R, and F1-score at both the pixel and object levels. Three representative disaster events from the xBD dataset, i.e., hurricane, tornado, and wildfire, were considered. As shown in Table~\ref{tab:localization_damage}, despite relying on automatically extracted building footprints, PUMA achieved pixel-level F1-scores of 84.11\%, 87.96\%, and 84.90\% for the hurricane, tornado, and wildfire events, respectively. The corresponding object-level F1-scores reached 80.02\%, 80.00\%, and 82.32\%, respectively. These results demonstrate that PUMA can be effectively integrated with automatic building localization to provide reliable building damage mapping when historical building footprints are unavailable, thereby extending its applicability to more practical post-disaster scenarios.

\subsubsection{Hyperparameter analysis}

A series of experiments were conducted to systematically evaluate the sensitivity of PUMA to its hyperparameters, with the results presented in Figure~\ref{fig:hyperparameters}. Here, $\gamma$ denotes the hyperparameter introduced in SAT, while $w_{e}$ denotes the weight of the entropy regularization term. Overall, PUMA exhibits stable performance across different settings of $k$, $w_{rep}$, $\alpha$, and $\gamma$, indicating limited sensitivity to these hyperparameters. A smaller $\beta$ accelerates the pseudo-label updating, whereas a moderate $w_{e}$ facilitates effective weakly supervised PEFT of the GFM.

\subsection{Real-world applications at the local scale}

\begin{table}[!b]
\centering
\caption{Building damage mapping results for the Eaton and Palisades subareas during the 2025 Los Angeles wildfires. Naive Supervision treats all unlabeled samples as negative; Standard Supervision additionally incorporates labeled undamaged buildings in equal number to the labeled damaged buildings; and Full Supervision assigns ground-truth labels to all unlabeled samples.}
\label{tab:local_area}
\resizebox{0.98\linewidth}{!}{%
\begin{tblr}{
  row{1} = {c},
  row{2} = {c},
  cell{1}{1} = {r=2}{},
  cell{1}{2} = {c=4}{},
  cell{1}{6} = {c=4}{},
  cell{3}{1} = {c=9}{},
  cell{4}{2} = {c},
  cell{4}{3} = {c},
  cell{4}{4} = {c},
  cell{4}{5} = {c},
  cell{4}{6} = {c},
  cell{4}{7} = {c},
  cell{4}{8} = {c},
  cell{4}{9} = {c},
  cell{5}{2} = {c},
  cell{5}{3} = {c},
  cell{5}{4} = {c},
  cell{5}{5} = {c},
  cell{5}{6} = {c},
  cell{5}{7} = {c},
  cell{5}{8} = {c},
  cell{5}{9} = {c},
  cell{6}{2} = {c},
  cell{6}{3} = {c},
  cell{6}{4} = {c},
  cell{6}{5} = {c},
  cell{6}{6} = {c},
  cell{6}{7} = {c},
  cell{6}{8} = {c},
  cell{6}{9} = {c},
  cell{7}{2} = {c},
  cell{7}{3} = {c},
  cell{7}{4} = {c},
  cell{7}{5} = {c},
  cell{7}{6} = {c},
  cell{7}{7} = {c},
  cell{7}{8} = {c},
  cell{7}{9} = {c},
  cell{8}{1} = {c=9}{},
  cell{9}{2} = {c},
  cell{9}{3} = {c},
  cell{9}{4} = {c},
  cell{9}{5} = {c},
  cell{9}{6} = {c},
  cell{9}{7} = {c},
  cell{9}{8} = {c},
  cell{9}{9} = {c},
  cell{10}{2} = {c},
  cell{10}{3} = {c},
  cell{10}{4} = {c},
  cell{10}{5} = {c},
  cell{10}{6} = {c},
  cell{10}{7} = {c},
  cell{10}{8} = {c},
  cell{10}{9} = {c},
  cell{11}{2} = {c},
  cell{11}{3} = {c},
  cell{11}{4} = {c},
  cell{11}{5} = {c},
  cell{11}{6} = {c},
  cell{11}{7} = {c},
  cell{11}{8} = {c},
  cell{11}{9} = {c},
  cell{12}{2} = {c},
  cell{12}{3} = {c},
  cell{12}{4} = {c},
  cell{12}{5} = {c},
  cell{12}{6} = {c},
  cell{12}{7} = {c},
  cell{12}{8} = {c},
  cell{12}{9} = {c},
  hline{1,3-4,8-9,13} = {-}{},
  hline{2} = {2-9}{},
}
Method                                                                                 & Eaton Subarea                &                              &                              &                              & Palisades Subarea            &                              &                              &                              \\
                                                                                       & OA                           & F1                           & P                            & R                            & OA                           & F1                           & P                            & R                            \\
$|\mathcal{D}_p|=50$ damaged buildings   &                              &                              &                              &                              &                              &                              &                              &                              \\
Naive Supervision                                                                               & 94.03\textsuperscript{±0.76} & 76.41\textsuperscript{±3.69} & 90.11\textsuperscript{±0.92} & 66.40\textsuperscript{±5.02} & 50.63\textsuperscript{±2.23} & 38.46\textsuperscript{±4.26} & 98.70\textsuperscript{±0.93} & 23.93\textsuperscript{±3.28} \\
Standard Supervision                                                                            & 91.73\textsuperscript{±0.74} & 77.54\textsuperscript{±1.47} & 64.34\textsuperscript{±2.20} & 97.63\textsuperscript{±0.75} & 92.87\textsuperscript{±1.08} & 94.26\textsuperscript{±0.93} & 98.26\textsuperscript{±0.29} & 90.58\textsuperscript{±1.80} \\
PUMA                                                                                   & 97.81\textsuperscript{±0.28} & 92.81\textsuperscript{±0.86} & 89.07\textsuperscript{±1.53} & 96.90\textsuperscript{±1.03} & 92.48\textsuperscript{±1.47} & 93.89\textsuperscript{±1.24} & 98.79\textsuperscript{±0.46} & 89.46\textsuperscript{±1.88} \\
Full Supervision                                                                                 & 97.60\textsuperscript{±0.62} & 92.30\textsuperscript{±1.85} & 87.16\textsuperscript{±3.12} & 98.11\textsuperscript{±0.34} & 97.44\textsuperscript{±0.04} & 98.01\textsuperscript{±0.03} & 98.26\textsuperscript{±0.42} & 97.77\textsuperscript{±0.44} \\
$|\mathcal{D}_p|=300$ damaged buildings &                              &                              &                              &                              &                              &                              &                              &                              \\
Naive Supervision                                                                               & 96.81\textsuperscript{±0.32} & 88.21\textsuperscript{±1.40} & 95.76\textsuperscript{±1.15} & 81.82\textsuperscript{±3.03} & 54.22\textsuperscript{±3.05} & 45.21\textsuperscript{±5.75} & 99.36\textsuperscript{±0.24} & 29.38\textsuperscript{±4.82} \\
Standard Supervision                                                                            & 96.53\textsuperscript{±0.64} & 89.25\textsuperscript{±1.77} & 81.56\textsuperscript{±3.06} & 98.61\textsuperscript{±0.10} & 97.34\textsuperscript{±0.21} & 97.93\textsuperscript{±0.17} & 98.87\textsuperscript{±0.08} & 96.99\textsuperscript{±0.27} \\
PUMA                                                                                   & 99.03\textsuperscript{±0.27} & 96.73\textsuperscript{±0.88} & 95.97\textsuperscript{±1.45} & 97.50\textsuperscript{±0.30} & 97.08\textsuperscript{±0.11} & 97.71\textsuperscript{±0.09} & 99.42\textsuperscript{±0.13} & 96.05\textsuperscript{±0.23} \\
Full Supervision                                                                                 & 98.93\textsuperscript{±0.08} & 96.42\textsuperscript{±0.28} & 94.65\textsuperscript{±0.52} & 98.26\textsuperscript{±0.04} & 98.09\textsuperscript{±0.19} & 98.52\textsuperscript{±0.14} & 98.74\textsuperscript{±0.14} & 98.30\textsuperscript{±0.15} 
\end{tblr}
}
\end{table}

\begin{figure}[!t]
\centering
\includegraphics[width=0.98\linewidth]{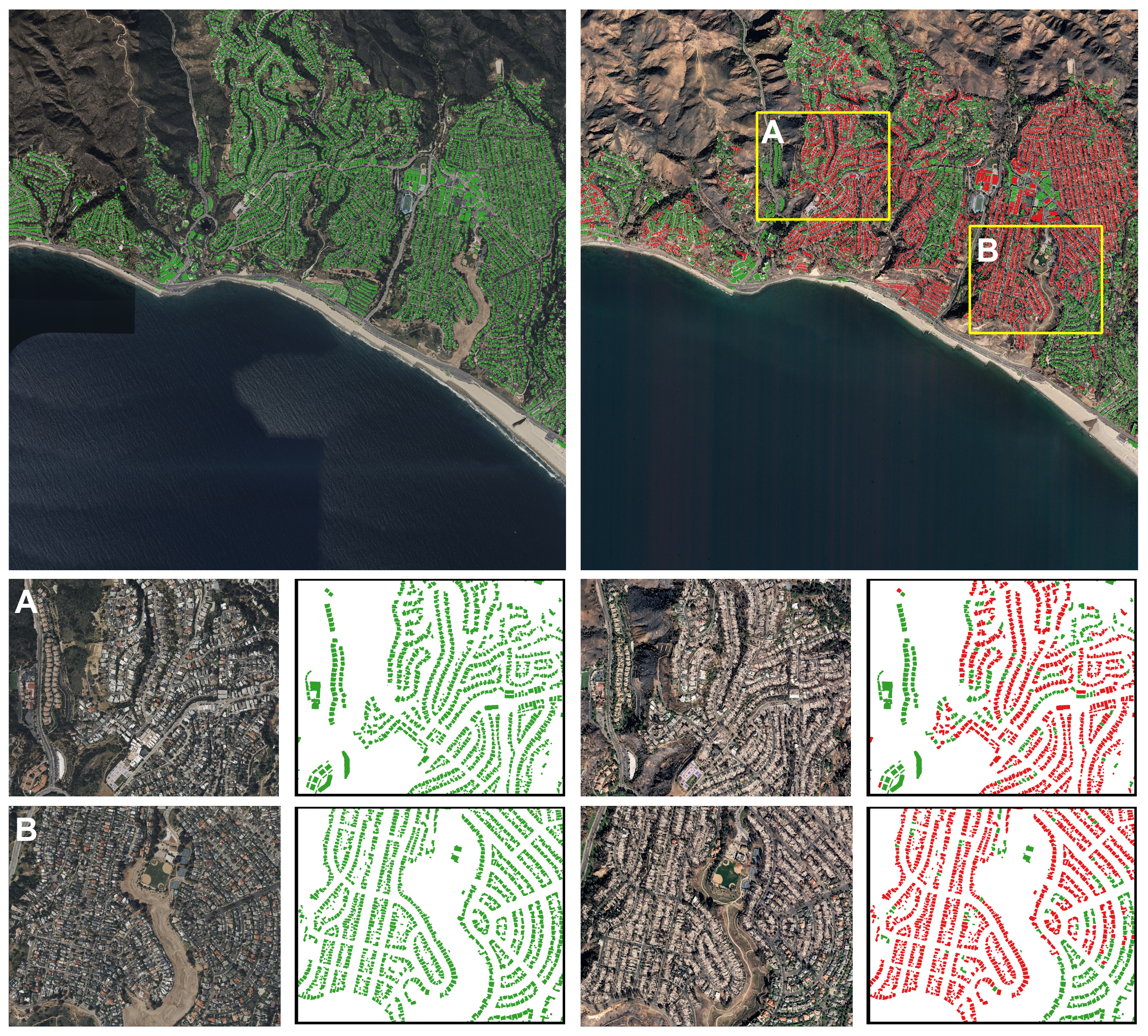}
\caption{Building damage mapping in the Palisades subarea of the Los Angeles wildfires. The left column shows pre-disaster imagery with building footprints, and the right column shows post-disaster imagery with the corresponding damage mapping results. Damaged and undamaged buildings are shown in \textcolor{red}{red} and \textcolor{darkgreen}{green}, respectively.}
\label{fig:local_wildfire_2}
\end{figure}

To demonstrate the effectiveness of the PUMA framework in real-world disaster scenarios, building damage mapping was conducted for the 2025 Los Angeles wildfires. The Eaton and Palisades fires were among the most destructive wildfire events in California's history. Three representative methods were compared to evaluate the performance of PUMA for GFM fine-tuning in real-world applications: Naive Supervision, which directly treats all unlabeled samples as negative; Standard Supervision, which additionally labels an equal number of undamaged buildings to match the number of labeled damaged buildings; and Full Supervision, which assigns ground-truth labels to all unlabeled samples. For all three methods, the GFM was fine-tuned using the cross-entropy loss. The quantitative results are presented in Table~\ref{tab:local_area}, while the corresponding building damage maps are shown in Figure~\ref{fig:local_wildfire_1} and Figure~\ref{fig:local_wildfire_2}.

\begin{figure}[!t]
\centering
\begin{subfigure}[b]{0.45\textwidth}
    \centering
    \includegraphics[width=\linewidth]{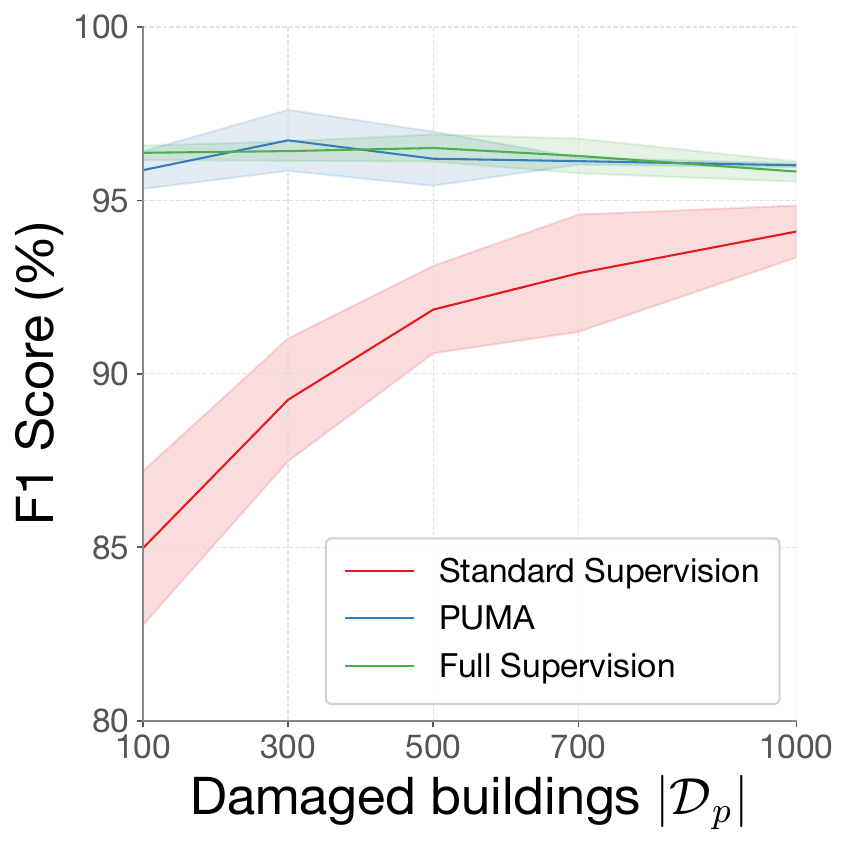}
    \caption{Label Efficiency under Limited Damaged Annotations}
    \label{fig:sample_size_1}
\end{subfigure}
\hspace{0.03\linewidth}
\begin{subfigure}[b]{0.45\textwidth}
    \centering
    \includegraphics[width=\linewidth]{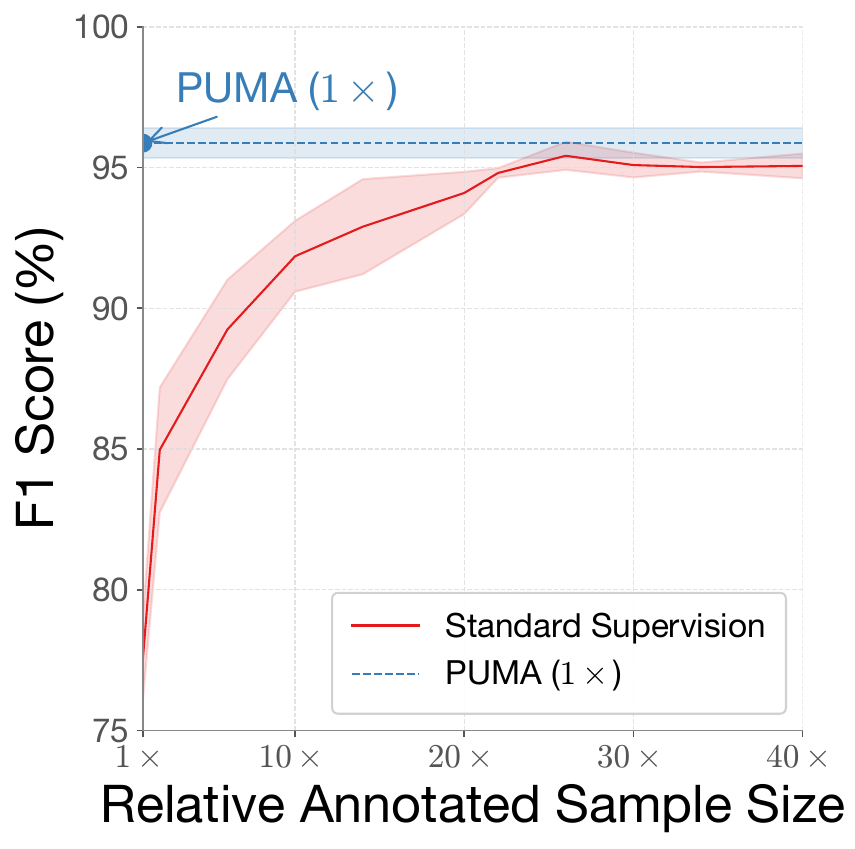}
    \caption{Label Efficiency Relative to Standard Supervision}
    \label{fig:sample_size_2}
\end{subfigure}
\caption{Label efficiency of PUMA compared with supervised PEFT methods in the Eaton subarea of the Los Angeles wildfires. A fixed set of 4,000 unlabeled buildings is used in all experiments, and $1\times$ corresponds to 100 annotated samples. (a) F1 scores under varying numbers of labeled damaged buildings. PUMA maintains performance close to Full Supervision and consistently outperforms Standard Supervision across annotation budgets. (b) PUMA requires only $1/40$ of the annotated samples used by Standard Supervision.}
\label{fig:sample_size_analysis}
\end{figure}

As indicated by the Naive Supervision results in Table~\ref{tab:local_area}, simply treating unlabeled samples as undamaged buildings is ineffective for real-world building damage mapping, as it leads to substantially reduced recall for damaged buildings. In contrast, PUMA achieved promising performance using only a small number of labeled damaged buildings (50 samples). Figure~\ref{fig:sample_size_analysis} further examines its label efficiency under different annotation budgets in the Eaton subarea of the Los Angeles wildfires. PUMA maintained consistently high F1 scores as the number of labeled damaged buildings varied, with performance remaining close to that of Full Supervision. In comparison, Standard Supervision was substantially more dependent on the availability of annotated samples and required a much larger annotation budget to approach PUMA's performance. The mapping results show that, within the image-covered area of Eaton, 3,752 of the 26,280 buildings were identified as damaged, corresponding to a damage rate of 14.28\%. In the image-covered area of Palisades, 5,073 of the 7,799 buildings were identified as damaged, corresponding to a damage rate of 65.05\%.

\section{Conclusion}\label{sec:conclusion}

The effective adaptation of GFMs for post-disaster assessment has been hindered by their limited capacity to learn damage-discriminative representations under the weak supervision required for rapid response. To address this challenge, PUMA integrates NeWER and LaTTE to enable iterative refinement of both representations and supervision signals. The two components mutually reinforce each other, and their interaction can be theoretically interpreted within a GEM-inspired framework. Extensive experiments across multiple disaster events demonstrated the effectiveness, robustness, and practical potential of PUMA. Its application to the 2025 Los Angeles wildfires further demonstrated the capability of the framework to accurately map building damage with a limited number of annotations. Overall, these findings highlight the potential of PUMA for label-efficient adaptation of GFMs to real-world applications. Future work will extend the framework to more complex damage assessment scenarios and broader weakly supervised remote sensing applications.

\section*{CRediT authorship contribution statement}
\textbf{Hengwei Zhao:} Writing - original draft, Visualization, Software, Methodology, Data curation, Conceptualization, Formal analysis, Investigation. \textbf{Qirui Wang:} Writing - review and editing, Data curation. \textbf{Zhuo Zheng:} Writing - review and editing. \textbf{Xiaoyan Lu:} Writing - review and editing. \textbf{Rusty Feagin:} Writing - review and editing. \textbf{Wenzhe Jiao:} Writing - review and editing, Supervision, Resources, Funding acquisition.

\section*{Declaration of competing interest}
The authors declare that they have no known competing financial interests or personal relationships that could have appeared to influence the work reported in this paper.

\section*{Acknowledgments}
We acknowledge support from Agriculture and Food Research Initiative program (2024-68016-43058) of the United States Department of Agriculture National Institute of Food and Agriculture. Portions of this research were conducted with the advanced computing resources provided by Texas A\&M High Performance Research Computing.

\bibliographystyle{elsarticle-harv}
\bibliography{ref}

\newpage
\appendix

\section{Derivation of the Gradient Magnitudes}\label{apx:gradient}
Because $\sg(\cdot)$ denotes the stop-gradient operator, $\bm{\tilde{z}}_{j}$ is treated as a constant when differentiating the losses with respect to $\bm{p}_{i}$. Let $\bm{\tilde{p}}_{i}=\bm{p}_{i}/\norm{\bm{p}_{i}}$ and $\bm{\tilde{z}}_{j}=\bm{z}_{j}/\norm{\bm{z}_{j}}$. The Jacobian of the $\ell_{2}$-normalization operation is given by
\begin{equation}
  \frac{\partial \tilde{\bm{p}}_{i}}{\partial \bm{p}_{i}} = \frac{1}{\norm{\bm{p}_{i}}}(\mathbf{I}-\tilde{\bm{p}}_{i}\tilde{\bm{p}}_{i}^{\top}).
\end{equation}
Consequently, the gradient of $\tilde{\bm{p}}_{i}^\top \tilde{\bm{z}}_{j}$ with respect to $\bm{p}_{i}$ is given by
\begin{equation}
  \begin{aligned}
    \frac{\partial \tilde{\bm{p}}_{i}^\top \tilde{\bm{z}}_{j}}{\partial \bm{p}_{i}} &= \frac{\partial \tilde{\bm{p}}_{i}}{\partial \bm{p}_{i}}{\tilde{\bm{z}}_{j}}\\
    &=\frac{1}{\norm{\bm{p}_{i}}}(\mathbf{I}-\tilde{\bm{p}}_{i}\tilde{\bm{p}}_{i}^{\top}){\tilde{\bm{z}}_{j}}\\
    &=\frac{1}{\norm{\bm{p}_{i}}}({\tilde{\bm{z}}_{j}}-\tilde{\bm{p}}_{i}\tilde{\bm{p}}_{i}^{\top}{\tilde{\bm{z}}_{j}}).
  \end{aligned}
\end{equation}
The squared norm of this gradient can be expressed as
\begin{equation}
  \begin{aligned}
    \norm{\frac{\partial \tilde{\bm{p}}_{i}^\top \tilde{\bm{z}}_{j}}{\partial \bm{p}_{i}}}^2 &= \frac{1}{\norm{\bm{p}_{i}}^2}\norm{{\tilde{\bm{z}}_{j}}-\tilde{\bm{p}}_{i}\tilde{\bm{p}}_{i}^{\top}{\tilde{\bm{z}}_{j}}}^2\\
    &= \frac{1-(\tilde{\bm{p}}_{i}^\top \tilde{\bm{z}}_{j})^2}{\norm{\bm{p}_{i}}^2},
  \end{aligned}
\end{equation}
where we have used $\norm{\tilde{\bm{p}}_{i}}=\norm{\tilde{\bm{z}}_{j}}=1$. Therefore,
\begin{equation}
  \norm{\frac{\partial \tilde{\bm{p}}_{i}^\top \tilde{\bm{z}}_{j}}{\partial \bm{p}_{i}}}=\frac{\sqrt{1-(\tilde{\bm{p}}_{i}^\top \tilde{\bm{z}}_{j})^2}}{\norm{\bm{p}_{i}}}.
\end{equation}

\noindent \textbf{Gradient and gradient magnitude of $\mathcal{L}_\text{rep}^\text{sup}(\bm{p}_{i},\sg(\bm{z}_{j}))$.} Consider the fully supervised representation loss
\begin{equation}
  \mathcal{L}_\text{rep}^\text{sup}(\bm{p}_{i},\sg(\bm{z}_j))=2(1-\bm{\tilde{p}}_{i}^{\top}\bm{\tilde{z}}_{j})\mathbb{I}\{y_i = y_j\}.
\end{equation}
Applying the chain rule yields
\begin{equation}
  \begin{aligned}
    \frac{\partial \mathcal{L}_\text{rep}^\text{sup}(\bm{p}_{i},\sg(\bm{z}_j))}{\partial \bm{p}_{i}} &= {-2}{\mathbb{I}\{y_i = y_j\}}{\frac{\partial \bm{\tilde{p}}_{i}^{\top}\bm{\tilde{z}}_{j}}{\partial \bm{p}_{i}}}\\
    &= {-}{\frac{2\mathbb{I}\{y_i = y_j\}}{\norm{\bm{p}_{i}}}}{({\tilde{\bm{z}}_{j}}-\tilde{\bm{p}}_{i}\tilde{\bm{p}}_{i}^{\top}{\tilde{\bm{z}}_{j}})}.
  \end{aligned}
\end{equation}
The corresponding gradient magnitude is therefore
\begin{equation}
    \norm{\frac{\partial \mathcal{L}_\text{rep}^\text{sup}(\bm{p}_{i},\sg(\bm{z}_j))}{\partial \bm{p}_{i}}} = {\frac{2\mathbb{I}\{y_i = y_j\}}{\norm{\bm{p}_{i}}}}{\sqrt{1-(\tilde{\bm{p}}_{i}^{\top}{\tilde{\bm{z}}}_{j})^2}}.
\end{equation}

\noindent \textbf{Gradient and gradient magnitude of $\mathcal{L}_\text{rep}^\text{sqrt}(\bm{p}_{i},\sg(\bm{z}_{j}))$.} The $\mathcal{L}_\text{rep}^\text{sqrt}$ is defined as
\begin{equation}
  \mathcal{L}_\text{rep}^\text{sqrt}(\bm{p}_{i},\sg(\bm{z}_j))=2\sqrt{k(1-\bm{\tilde{p}}_{i}^{\top}\bm{\tilde{z}}_{j})}\mathbb{I}\{y_i = y_j\},
\end{equation}
where $k>0$ is a hyperparameter controlling how tightly the representations are clustered around the mean direction, described in Section~\ref{sec:EM}. For $\bm{\tilde{p}}_{i}^{\top}\bm{\tilde{z}}_{j} < 1$, its derivative with respect to $\bm{\tilde{p}}_{i}^{\top}\bm{\tilde{z}}_{j}$ is given by
\begin{equation}
  \begin{aligned}
    {\frac{\partial \mathcal{L}_\text{rep}^\text{sqrt}(\bm{p}_{i},\sg(\bm{z}_j))}{\partial \bm{\tilde{p}}_{i}^{\top}\bm{\tilde{z}}_{j}}} &= {2\mathbb{I}\{y_i = y_j\}}{\frac{-k}{2\sqrt{k(1-\bm{\tilde{p}}_{i}^{\top}\bm{\tilde{z}}_{j})}}}\\
    &= {-}{\frac{\sqrt{k}}{\sqrt{1-\bm{\tilde{p}}_{i}^{\top}\bm{\tilde{z}}_{j}}}}\mathbb{I}\{y_i = y_j\}.
  \end{aligned}
\end{equation}
Applying the chain rule yields
\begin{equation}
  \begin{aligned}
    {\frac{\partial \mathcal{L}_\text{rep}^\text{sqrt}(\bm{p}_{i},\sg(\bm{z}_j))}{\partial \bm{p}_{i}}} &= {\frac{\partial \mathcal{L}_\text{rep}^\text{sqrt}(\bm{p}_{i},\sg(\bm{z}_j))}{\partial \bm{\tilde{p}}_{i}^{\top}\bm{\tilde{z}}_{j}}}{\frac{\partial \bm{\tilde{p}}_{i}^{\top}\bm{\tilde{z}}_{j}}{\partial \bm{p}_{i}}}\\
    &= {-}{\frac{\sqrt{k}\mathbb{I}\{y_i = y_j\}}{\norm{\bm{p}_{i}}\sqrt{1-\bm{\tilde{p}}_{i}^{\top}\bm{\tilde{z}}_{j}}}}({\tilde{\bm{z}}_{j}}-\tilde{\bm{p}}_{i}\tilde{\bm{p}}_{i}^{\top}{\tilde{\bm{z}}_{j}}).
  \end{aligned}
\end{equation}
The corresponding gradient magnitude is
\begin{equation}
  \begin{aligned}
    \norm{\frac{\partial \mathcal{L}_\text{rep}^\text{sqrt}(\bm{p}_{i},\sg(\bm{z}_j))}{\partial \bm{p}_{i}}} &= {\frac{\sqrt{k}\mathbb{I}\{y_i = y_j\}}{\norm{\bm{p}_{i}}\sqrt{1-\bm{\tilde{p}}_{i}^{\top}\bm{\tilde{z}}_{j}}}}{\sqrt{1-(\bm{\tilde{p}}_{i}^{\top}\bm{\tilde{z}}_{j})^{2}}}\\
    &= {\frac{\sqrt{k}\mathbb{I}\{y_i = y_j\}}{\norm{\bm{p}_{i}}}}{\sqrt{\frac{(1-\bm{\tilde{p}}_{i}^{\top}\bm{\tilde{z}}_{j})(1+\bm{\tilde{p}}_{i}^{\top}\bm{\tilde{z}}_{j})}{1-\bm{\tilde{p}}_{i}^{\top}\bm{\tilde{z}}_{j}}}}\\
    &= {\frac{\sqrt{k}\mathbb{I}\{y_i = y_j\}}{\norm{\bm{p}_{i}}}}{\sqrt{1+\bm{\tilde{p}}_{i}^{\top}\bm{\tilde{z}}_{j}}}.
  \end{aligned}
\end{equation}
Therefore,
\begin{equation}
    \norm{\frac{\partial \mathcal{L}_\text{rep}^\text{sqrt}(\bm{p}_{i},\sg(\bm{z}_j))}{\partial \bm{p}_{i}}} = {\frac{\sqrt{k}\mathbb{I}\{y_i = y_j\}}{\norm{\bm{p}_{i}}}}{\sqrt{1+\bm{\tilde{p}}_{i}^{\top}\bm{\tilde{z}}_{j}}}.
\end{equation}
When $\mathbb{I}\{y_i = y_j\}=0$, both losses are identically zero, and hence their gradients with respect to $\bm{p}_{i}$ are also zero. It should also be noted that $\mathcal{L}_\text{rep}^\text{sqrt}(\bm{p}_{i},\sg(\bm{z}_j))$ is not differentiable at $\bm{\tilde{p}}_{i}^{\top}\bm{\tilde{z}}_{j}=1$, corresponding to $\bm{\tilde{p}}_{i} = \bm{\tilde{z}}_{j}$. In practice, numerical stability is ensured through the combination of the prediction head and element-wise clipping, with each element of the embeddings $\tilde{\bm{p}}_{i}$ and $\tilde{\bm{z}}_{j}$ constrained to the interval $[10^{-4},1-10^{-4}]$.

\noindent \textbf{Gradient and gradient magnitude of $\mathcal{L}_\text{rep}^\text{log}(\bm{p}_{i},\sg(\bm{z}_{j}))$.} The $\mathcal{L}_\text{rep}^\text{log}$ is defined as
\begin{equation}
  \mathcal{L}_\text{rep}^\text{log}(\bm{p}_{i},\sg(\bm{z}_j))=2\log{(1-\bm{\tilde{p}}_{i}^{\top}\bm{\tilde{z}}_{j})}\mathbb{I}\{y_i = y_j\}.
\end{equation}
For $\bm{\tilde{p}}_{i}^{\top}\bm{\tilde{z}}_{j} < 1$, its derivative with respect to $\bm{\tilde{p}}_{i}^{\top}\bm{\tilde{z}}_{j}$ is given by
\begin{equation}
  \begin{aligned}
    {\frac{\partial \mathcal{L}_\text{rep}^\text{log}(\bm{p}_{i},\sg(\bm{z}_j))}{\partial \bm{\tilde{p}}_{i}^{\top}\bm{\tilde{z}}_{j}}} 
    &= {-}{\frac{2}{1-\bm{\tilde{p}}_{i}^{\top}\bm{\tilde{z}}_{j}}}\mathbb{I}\{y_i = y_j\}.
  \end{aligned}
\end{equation}
Applying the chain rule yields
\begin{equation}
  \begin{aligned}
    {\frac{\partial \mathcal{L}_\text{rep}^\text{log}(\bm{p}_{i},\sg(\bm{z}_j))}{\partial \bm{p}_{i}}} &= {\frac{\partial \mathcal{L}_\text{rep}^\text{log}(\bm{p}_{i},\sg(\bm{z}_j))}{\partial \bm{\tilde{p}}_{i}^{\top}\bm{\tilde{z}}_{j}}}{\frac{\partial \bm{\tilde{p}}_{i}^{\top}\bm{\tilde{z}}_{j}}{\partial \bm{p}_{i}}}\\
    &= {-}{\frac{2\mathbb{I}\{y_i = y_j\}}{\norm{\bm{p}_{i}}(1-\bm{\tilde{p}}_{i}^{\top}\bm{\tilde{z}}_{j})}}({\tilde{\bm{z}}_{j}}-\tilde{\bm{p}}_{i}\tilde{\bm{p}}_{i}^{\top}{\tilde{\bm{z}}_{j}}).
  \end{aligned}
\end{equation}
The corresponding gradient magnitude is
\begin{equation}
  \begin{aligned}
    \norm{\frac{\partial \mathcal{L}_\text{rep}^\text{log}(\bm{p}_{i},\sg(\bm{z}_j))}{\partial \bm{p}_{i}}} &= {\frac{2\mathbb{I}\{y_i = y_j\}}{\norm{\bm{p}_{i}}(1-\bm{\tilde{p}}_{i}^{\top}\bm{\tilde{z}}_{j})}}{\sqrt{1-(\bm{\tilde{p}}_{i}^{\top}\bm{\tilde{z}}_{j})^{2}}}\\
    &= {\frac{2\mathbb{I}\{y_i = y_j\}}{\norm{\bm{p}_{i}}}}{\sqrt{\frac{(1-\bm{\tilde{p}}_{i}^{\top}\bm{\tilde{z}}_{j})(1+\bm{\tilde{p}}_{i}^{\top}\bm{\tilde{z}}_{j})}{(1-\bm{\tilde{p}}_{i}^{\top}\bm{\tilde{z}}_{j})^2}}}\\
    &= {\frac{2\mathbb{I}\{y_i = y_j\}}{\norm{\bm{p}_{i}}}}{\sqrt{\frac{1+\bm{\tilde{p}}_{i}^{\top}\bm{\tilde{z}}_{j}}{1-\bm{\tilde{p}}_{i}^{\top}\bm{\tilde{z}}_{j}}}}.
  \end{aligned}
\end{equation}
Therefore,
\begin{equation}
    \norm{\frac{\partial \mathcal{L}_\text{rep}^\text{log}(\bm{p}_{i},\sg(\bm{z}_j))}{\partial \bm{p}_{i}}} = {\frac{2\mathbb{I}\{y_i = y_j\}}{\norm{\bm{p}_{i}}}}{\sqrt{\frac{1+\bm{\tilde{p}}_{i}^{\top}\bm{\tilde{z}}_{j}}{1-\bm{\tilde{p}}_{i}^{\top}\bm{\tilde{z}}_{j}}}}.
\end{equation}
In practice, as with $\mathcal{L}_\text{rep}^\text{sqrt}(\bm{p}_{i},\sg(\bm{z}_j))$, numerical stability is ensured by combining the prediction head with element-wise clipping, which constrains each element of the embeddings $\tilde{\bm{p}}_{i}$ and $\tilde{\bm{z}}_{j}$ to the interval $[10^{-4},1-10^{-4}]$.

\section{Theoretical Analysis from a GEM Perspective}\label{apx:EM}
\subsection{Overview}
To establish a theoretical foundation for PUMA, this paper interprets its alternating GFM fine-tuning procedure from the perspective of GEM when $\mathcal{L}_\text{rep}^\text{sqrt}$ is employed as the representation refinement loss. By parameterizing the latent posterior using classifier predictions, the variational lower bound on the unlabeled-data log-likelihood can be related to the representation refinement loss under mild assumptions. Since the labeled damaged samples have reliable annotations, the following analysis focuses on the unlabeled data.

Let $\bm{\theta}$ denote the trainable parameters of the GFM and its associated trainable components. The marginal log-likelihood of the unlabeled data is given by
\begin{equation}
\mathcal{J}(\bm{\theta})
=\sum_{\bm{x}_{i} \in \mathcal{D}_{u}}\log p(\bm{x}_{i};\bm{\theta})
=\sum_{\bm{x}_{i} \in \mathcal{D}_{u}}\log \sum_{y_{i} \in \mathcal{Y}}p(\bm{x}_{i}, y_{i};\bm{\theta}),
\label{eq:unlabeled_loglihood}
\end{equation}
where $y_{i} \in \mathcal{Y}=\{0,1\}$ denotes the latent class label associated with $\bm{x}_{i}$.
Directly maximizing Eq.~(\ref{eq:unlabeled_loglihood}) is challenging because the class labels of the unlabeled samples are unobserved. Therefore, a variational distribution is introduced to derive a tractable lower bound.

\subsection{Latent Posterior Inference: E-Step}
Let $\mathcal{Q}(y_{i})$ denote a variational distribution over the latent class label $y_{i}$. By introducing $\mathcal{Q}(y_{i})$ and applying Jensen's inequality, the marginal log-likelihood admits the following evidence lower bound (ELBO):
\begin{equation}
    \begin{aligned}
        \mathcal{J}(\bm{\theta})
        &=\sum_{\bm{x}_{i} \in \mathcal{D}_{u}}\log \sum_{y_{i} \in \mathcal{Y}}\mathcal{Q}(y_{i})\frac{p(\bm{x}_{i}, y_{i};\bm{\theta})}{\mathcal{Q}(y_{i})}\\
        &\ge \sum_{\bm{x}_{i} \in \mathcal{D}_{u}} \sum_{y_{i} \in \mathcal{Y}}\mathcal{Q}(y_{i})\log \frac{p(\bm{x}_{i}, y_{i};\bm{\theta})}{\mathcal{Q}(y_{i})}\\
        &\coloneqq \mathrm{ELBO}(\bm{\theta},\mathcal{Q}).
    \end{aligned}
\end{equation}
The gap between the marginal log-likelihood and the ELBO can be expressed as
\begin{equation}
    \begin{aligned}
        \mathcal{J}(\bm{\theta})-\mathrm{ELBO}(\bm{\theta},\mathcal{Q})&=\sum_{\bm{x}_{i} \in \mathcal{D}_{u}}\sum_{y_{i} \in \mathcal{Y}}\mathcal{Q}(y_{i})\log \frac{p(\bm{x}_{i};\bm{\theta})\mathcal{Q}(y_{i})}{p(\bm{x}_{i}, y_{i};\bm{\theta})}\\
        &=\sum_{\bm{x}_{i} \in \mathcal{D}_{u}}\sum_{y_{i} \in \mathcal{Y}}\mathcal{Q}(y_{i})\log \frac{\mathcal{Q}(y_{i})}{p(y_{i} \mid \bm{x}_{i};\bm{\theta})}\\
        &=\sum_{\bm{x}_{i} \in \mathcal{D}_{u}}\mathrm{KL}\left(\mathcal{Q}(y_{i}) \parallel p(y_{i} \mid \bm{x}_{i};\bm{\theta})\right).
    \end{aligned}
\end{equation}
Therefore, the ELBO becomes tight when the variational distribution matches the exact posterior:
\begin{equation}
    \mathcal{Q}(y_{i})=p(y_{i} \mid \bm{x}_{i};\bm{\theta}).
\end{equation}
For simplicity, the E-step is approximated using a hard-assignment strategy: $\mathcal{Q}(y_i)=\mathbb{I}\{\tilde{y}_{i}=y_{i}\}$, where $\tilde{y}_{i}=\mathop{\arg\max}f(\bm{r}_{i})$ denotes the predicted label of $\bm{x}_{i}$. Accordingly, the subsequent parameter update optimizes the following objective:
\begin{equation}
    \label{eq:final_log_likelihood}
    \bm{\theta}^{*}=\mathop{\arg\max}\limits_{\bm{\theta}}\sum_{\bm{x}_{i} \in \mathcal{D}_{u}} \sum_{y_{i} \in \mathcal{Y}}\mathbb{I}\{\tilde{y}_{i}=y_{i}\}\log p(\bm{x}_{i}, y_{i};\bm{\theta}).
\end{equation}

\subsection{Log-Likelihood Maximization: M-step}
Minimizing the proposed representation refinement loss $\mathcal{L}_\text{rep}^\text{sqrt}$ can be interpreted as maximizing a lower bound on the log-likelihood of the unlabeled data.

\subsubsection{Class-Conditional Objective Decomposition}
For analytical convenience, all unlabeled samples are considered at each iteration and partitioned into class-specific subsets $\mathcal{U}_{c}=\{ \bm{x}_{i} \in \mathcal{D}_{u} \mid \tilde{y}_{i}=c \}$:
\begin{equation}
    \begin{aligned}
        &\mathop{\arg\max}\limits_{\bm{\theta}}\sum_{\bm{x}_{i} \in \mathcal{D}_{u}}                                                                                                                       \sum_{y_{i} \in \mathcal{Y}}\mathbb{I}\{\tilde{y}_{i}=y_{i}\}\log p(\bm{x}_{i}, y_{i};\bm{\theta})\\
        =&\mathop{\arg\max}\limits_{\bm{\theta}}\sum_{\bm{x}_{i} \in \mathcal{D}_{u}}                                                                                                                       \sum_{y_{i} \in \mathcal{Y}}\mathbb{I}\{\tilde{y}_{i}=y_{i}\} \left[ \log p(\bm{x}_{i} \mid y_{i};\bm{\theta})+ \log p(y_{i}) \right]\\
        =&\mathop{\arg\max}\limits_{\bm{\theta}}\sum_{\mathcal{U}_{c} \in \mathcal{D}_{u}} \sum_{\bm{x}_{i} \in \mathcal{U}_c}\log p(\bm{x}_{i} \mid y_{i}=c;\bm{\theta}),
    \end{aligned}
\end{equation}
where the hard assignments are fixed and $p(y_{i})$ is independent of the model parameters $\bm{\theta}$ during the M-step.

To facilitate the analysis, the augmentation operation is omitted, the prediction head is treated as an identity mapping, and $\bm{\tilde{p}}_{i}=t(\bm{x}_{i})$. For each predicted class $c$, the empirical class center is defined as
\begin{equation}
    \bm{\mu}_{c}=\frac{1}{\lvert \mathcal{U}_{c} \rvert}\sum_{\bm{x}_{i} \in \mathcal{U}_{c}} t(\bm{x}_{i}).
\end{equation}

Assume that samples assigned to the same class in the non-contrastive output space follow a $d$-variate von Mises-Fisher distribution: $\varphi (\bm{x}_{i}\mid \tilde{\bm{\mu}}_{c},\kappa)=C_{d}(\kappa)e^{\kappa\bm{\tilde{\mu}}_{c}^{\top}t(\bm{x}_{i})}$, where $\bm{\tilde{\mu}}_{c}=\bm{\mu}_{c}/\norm{\bm{\mu}_{c}}$, $\kappa > 0$ is a fixed concentration parameter, and $C_{d}(\kappa)$ is the normalization constant. Under this assumption, we obtain
\begin{equation}
    \begin{aligned}
        &\mathop{\arg\max}\limits_{\bm{\theta}}\sum_{\mathcal{U}_{c} \in \mathcal{D}_{u}} \sum_{\bm{x}_{i} \in \mathcal{U}_c}\log p(\bm{x}_{i} \mid y_{i}=c;\bm{\theta})\\
        =&\mathop{\arg\max}\limits_{\bm{\theta}}\sum_{\mathcal{U}_{c} \in \mathcal{D}_{u}} \sum_{\bm{x}_{i} \in \mathcal{U}_c}\kappa \tilde{\bm{\mu}}_{c}^{\top}t(\bm{x}_{i})\\
        =&\mathop{\arg\max}\limits_{\bm{\theta}}\sum_{\mathcal{U}_{c} \in \mathcal{D}_{u}} \kappa {\lvert \mathcal{U}_{c} \rvert}\norm{\bm{\mu}_{c}}.
    \end{aligned}
\end{equation}

\subsubsection{Bridging representation refinement loss and the decomposed objective}
For $\bm{x}_{i},\bm{x}_{j} \in \mathcal{U}_{c}$, the representation refinement loss $\mathcal{L}_\text{rep}^\text{sqrt}$ for a same-class pair is given by
\begin{equation}
    \mathcal{L}_\text{rep}^\text{sqrt}(\bm{p}_{i},\sg(\bm{z}_{j}))=2\sqrt{k\left(1-\tilde{\bm{p}}_{i}^{\top}\tilde{\bm{z}}_{j}\right)}.
\end{equation}
It then follows that
\begin{equation}
    \begin{aligned}
        \tilde{\mathcal{R}}(\bm{x})&=\frac{1}{n_{u}}\sum_{\bm{x}_{i} \in \mathcal{D}_{u}}\left\{\frac{2}{\lvert \mathcal{Z}_{i} \rvert} \sum_{\bm{z}_{+} \in \mathcal{Z}_{i}} \sqrt{k(1-\tilde{\bm{p}}_{i}^{\top}\tilde{\bm{z}}_{+})} \right\}\\
        &=\frac{\sqrt{2k}}{n_{u}}\sum_{\bm{x}_{i} \in \mathcal{D}_{u}}\frac{1}{\lvert \mathcal{Z}_{i} \rvert}\sum_{\bm{z}_{+} \in \mathcal{Z}_{i}}\norm{\tilde{\bm{p}}_{i}-\tilde{\bm{z}}_{+}}\\
        &\approx \frac{\sqrt{2k}}{n_{u}}\sum_{\mathcal{U}_{c} \in \mathcal{D}_{u}}\sum_{\bm{x}_{i} \in \mathcal{U}_{c}}\frac{1}{\lvert \mathcal{U}_{c} \rvert}\sum_{\bm{x}_{j} \in \mathcal{U}_{c}}\norm{t(\bm{x}_{i})-t(\bm{x}_{j})}\\
        &= \frac{\sqrt{2k}}{n_{u}}\sum_{\mathcal{U}_{c} \in \mathcal{D}_{u}}\frac{1}{\lvert \mathcal{U}_{c} \rvert}\sum_{\bm{x}_{i},\bm{x}_{j} \in \mathcal{U}_{c}}\norm{t(\bm{x}_{i})-t(\bm{x}_{j})},
    \end{aligned}
\end{equation}
where $\mathcal{Z}_{i}$ denotes the positive peer set of $\bm{x}_{i}$. ${\lvert \mathcal{U}_{c} \rvert} \approx {\lvert \mathcal{U}_{c} \rvert}-1 ={\lvert \mathcal{Z}_{i} \rvert}$ since the number of training samples is usually large.

For a fixed class-specific subset $\mathcal{U}_{c}$,
\begin{equation}
    \begin{aligned}
        \norm{\bm{\mu}_{c}}^{2}
        &=\norm{\frac{1}{\lvert \mathcal{U}_{c} \rvert}\sum_{\bm{x}_{i} \in \mathcal{U}_{c}}t(\bm{x}_{i})}^2\\
        &=\frac{1}{{\lvert \mathcal{U}_{c} \rvert}^{2}}\sum_{\bm{x}_{i},\bm{x}_{j} \in \mathcal{U}_{c}}t(\bm{x}_{i})^{\top}t(\bm{x}_{j})\\
        &=\frac{1}{{\lvert \mathcal{U}_{c} \rvert}^{2}}\sum_{\bm{x}_{i},\bm{x}_{j} \in \mathcal{U}_{c}}\left( 1-\frac{1}{2}\norm{t(\bm{x}_{i})-t(\bm{x}_{j})}^2 \right)\\
        &=1-\frac{1}{{2\lvert \mathcal{U}_{c} \rvert}^{2}}\sum_{\bm{x}_{i},\bm{x}_{j} \in \mathcal{U}_{c}}\norm{t(\bm{x}_{i})-t(\bm{x}_{j})}^2.
    \end{aligned}
\end{equation}
Since the representations are unit vectors, the triangle inequality gives $0 \leq \norm{t(\bm{x}_{i})-t(\bm{x}_{j})} \leq 2$. Consequently,
\begin{equation}
    \norm{t(\bm{x}_{i})-t(\bm{x}_{j})}^2 \leq 2\norm{t(\bm{x}_{i})-t(\bm{x}_{j})}.
\end{equation}
Moreover, since $0 \leq \norm{\bm{\mu}_{c}} \leq 1$, we have
\begin{equation}
    \norm{\bm{\mu}_{c}} \geq \norm{\bm{\mu}_{c}}^{2} \geq 1-\frac{1}{{\lvert \mathcal{U}_{c} \rvert}^{2}}\sum_{\bm{x}_{i},\bm{x}_{j} \in \mathcal{U}_{c}}\norm{t(\bm{x}_{i})-t(\bm{x}_{j})}.
\end{equation}
Summing over all class-specific subsets yields
\begin{equation}
    \begin{aligned}
        \sum_{\mathcal{U}_{c} \in \mathcal{D}_{u}} \kappa {\lvert \mathcal{U}_{c} \rvert}\norm{\bm{\mu}_{c}}
        &\geq \kappa n_{u}-\sum_{\mathcal{U}_{c} \in \mathcal{D}_{u}}\frac{\kappa}{\lvert \mathcal{U}_{c} \rvert}\sum_{\bm{x}_{i},\bm{x}_{j} \in \mathcal{U}_{c}}\norm{t(\bm{x}_{i})-t(\bm{x}_{j})}\\
        &=\kappa n_{u}-n_{u}\tilde{\mathcal{R}}(\bm{x}),
    \end{aligned}
    \label{eq:lower_bound}
\end{equation}
where $k$ can be related to the concentration parameter $\kappa$ through $k=\kappa^2/2$. Hence, minimizing the proposed representation refinement loss $\mathcal{L}_\text{rep}^\text{sqrt}$ increases a lower bound on the log-likelihood of the unlabeled data.

\subsection{Characterization of the Bound Gap}

\subsubsection{Definition of the Bound Gap}
Define the gap between the $\sum_{\mathcal{U}_{c} \in \mathcal{D}_{u}} \kappa {\lvert \mathcal{U}_{c} \rvert}\norm{\bm{\mu}_{c}}$ and $\kappa n_{u}-n_{u}\tilde{\mathcal{R}}(\bm{x})$ as
\begin{equation}
    \begin{aligned}
        \Delta &= \sum_{\mathcal{U}_{c} \in \mathcal{D}_{u}} \kappa {\lvert \mathcal{U}_{c} \rvert}\norm{\bm{\mu}_{c}}-\kappa n_{u}+n_{u}\tilde{\mathcal{R}}(\bm{x})\\
        &=\sum_{\mathcal{U}_{c} \in \mathcal{D}_{u}}\kappa {\lvert \mathcal{U}_{c} \rvert}\norm{\bm{\mu}_{c}}-\kappa n_{u}+\sum_{\mathcal{U}_{c} \in \mathcal{D}_{u}}\frac{\kappa}{\lvert \mathcal{U}_{c} \rvert}\sum_{\bm{x}_{i},\bm{x}_{j} \in \mathcal{U}_{c}}\norm{t(\bm{x}_{i})-t(\bm{x}_{j})}\\
        &=\sum_{\mathcal{U}_{c} \in \mathcal{D}_{u}}\kappa {\lvert \mathcal{U}_{c} \rvert}\left( \norm{\bm{\mu}_{c}}-1+\frac{1}{{\lvert \mathcal{U}_{c} \rvert}^2}\sum_{\bm{x}_{i},\bm{x}_{j} \in \mathcal{U}_{c}}\norm{t(\bm{x}_{i})-t(\bm{x}_{j})} \right).
    \end{aligned}
\end{equation}
By adding and subtracting $\norm{\bm{\mu}_{c}}^2$, the gap can be further expressed as
\begin{equation}
    \begin{aligned}
        \Delta &= \sum_{\mathcal{U}_{c} \in \mathcal{D}_{u}}\kappa {\lvert \mathcal{U}_{c} \rvert}\left( \norm{\bm{\mu}_{c}}-\norm{\bm{\mu}_{c}}^2+\frac{1}{{\lvert \mathcal{U}_{c} \rvert}^2}\sum_{\bm{x}_{i},\bm{x}_{j} \in \mathcal{U}_{c}}d_{ij}-(1-\norm{\bm{\mu}_{c}}^2) \right)\\
        &= \sum_{\mathcal{U}_{c} \in \mathcal{D}_{u}}\kappa {\lvert \mathcal{U}_{c} \rvert}\left[ \norm{\bm{\mu}_{c}}(1-\norm{\bm{\mu}_{c}})+\frac{1}{2{\lvert \mathcal{U}_{c} \rvert}^2}\sum_{\bm{x}_{i},\bm{x}_{j} \in \mathcal{U}_{c}}d_{ij}(2-d_{ij}) \right],
    \end{aligned}
    \label{eq:gap}
\end{equation}
where $d_{ij}=\norm{t(\bm{x}_{i})-t(\bm{x}_{j})}$.

Since $0 \leq \norm{\bm{\mu}_{c}} \leq 1$ and $0 \leq d_{ij} \leq 2$, both terms inside the brackets are nonnegative. Therefore, $$\Delta \geq 0.$$

\subsubsection{Tightness Conditions}
Equation~(\ref{eq:lower_bound}) is tight if and only if $\Delta=0$. Since every term in Eq.~(\ref{eq:gap}) is nonnegative, equality requires that, for every nonempty subset $\mathcal{U}_{c}$,
\begin{equation}
    \norm{\bm{\mu}_{c}}(1-\norm{\bm{\mu}_{c}})=0
\end{equation}
and
\begin{equation}
    d_{ij}(2-d_{ij})=0, \qquad \forall \bm{x}_{i},\bm{x}_{j} \in \mathcal{U}_{c}.
\end{equation}
Therefore, $$\norm{\bm{\mu}_{c}} \in \left\{ 0,1 \right\}$$ and $$d_{ij} \in \left\{ 0,2 \right\}.$$
The practically relevant equality condition is
\begin{equation}
    d_{ij}=0,\qquad \norm{\bm{\mu}_{c}}=1.
\end{equation}
A second, degenerate equality case occurs when the representations within a subset are evenly distributed between two antipodal directions, resulting in $\norm{\bm{\mu}_{c}}=0$. Although this case satisfies the algebraic equality conditions, it does not correspond to the desired cluster-tightening behavior.

\subsubsection{Asymptotic Tightness}
The gap does not decrease monotonically at every optimization step. Nevertheless, it vanishes asymptotically as the $\mathcal{L}_\text{rep}^\text{sqrt}$ approaches zero.

Since every term in $\tilde{\mathcal{R}}(\bm{x})$ is nonnegative, $$\tilde{\mathcal{R}}(\bm{x}) \to 0$$ implies $$\sum_{\bm{x}_{i},\bm{x}_{j} \in \mathcal{U}_{c}}d_{ij} \to 0.$$ For a fixed finite class, the nonnegativity of $d_{ij}$ implies $$d_{ij} \to 0$$ for all within-class pairs. It then follows that $$\norm{\bm{\mu}_{c}} \to 1.$$ Substituting $\norm{\bm{\mu}_{c}} \to 1$ and $d_{ij} \to 0$ into Eq.~(\ref{eq:gap}) yields $$\Delta \to 0.$$ Therefore, although strict stepwise monotonic shrinkage of the gap is not guaranteed, the lower bound becomes asymptotically tight as the within-class representations become increasingly aligned.

\end{document}